\documentclass{article}

\usepackage[dblblindworkshop, final]{neurips_2026}

\usepackage[utf8]{inputenc} 
\usepackage[T1]{fontenc}    
\usepackage{hyperref}       
\usepackage{url}            
\usepackage{booktabs}       
\usepackage{amsmath}        
\usepackage{amsfonts}       
\usepackage{nicefrac}       
\usepackage{microtype}      
\usepackage[table]{xcolor}  
\usepackage{multirow}
\usepackage{booktabs}
\usepackage{tabularx}
\usepackage{array}
\usepackage{makecell}
\usepackage{wrapfig}
\usepackage{subcaption}
\usepackage{graphicx}

\newcommand{\stepG}{\mathcal{G}}   
\newcommand{\stepW}{\mathcal{W}}   
\newcommand{\stepR}{\mathcal{R}}   

\newcommand{\settingT}{\tau}        
\newcommand{\settingF}{f}           

\newcommand{\nextgqa}{\textsc{NExT-GQA}}
\newcommand{\cgbench}{\textsc{CG-Bench}}
\newcommand{\argus}{\textsc{Argus}}
\newcommand{\elvhalluc}{\textsc{ELV-Halluc}}
\newcommand{\lvbench}{\textsc{LVBench}}
\newcommand{\drvbench}{\textsc{Dr.V-Bench}}
\newcommand{\drv}{\textsc{Dr.V}}

\newcommand{\condfull}{\texttt{full}}
\newcommand{\condpred}{\texttt{predicted}}
\newcommand{\condoracle}{\texttt{oracle}}
\newcommand{\condrandom}{\texttt{random}}
\newcommand{\condoff}{\texttt{off-target}}

\newcommand{\videor}{\textsc{Video-R1}}
\newcommand{\videomind}{\textsc{VideoMind}}

\newcommand{\eva}{\textsc{EVA}}

\newcommand{\ArgusCostH}{\text{Cost}_\text{H}}
\newcommand{\ArgusCostO}{\text{Cost}_\text{O}}

\title{Beneath the Scores: Rethinking Hallucination Evaluation for Video Understanding Models}
\workshoptitle{TAE (Trust-AI-Eval): Can We Trust AI Evaluation?}

\author{%
  Shuzhi Gong \qquad Fengze Sun \qquad Yuansan Liu \\
  The University of Melbourne \\
  \texttt{\{shuzhi, fengze.sun1, yuansan.liu1\}@unimelb.edu.au}
}

\begin{document}

\maketitle

\begin{abstract}
Video understanding is increasingly performed by multi-stage LLM agents that separate temporal grounding, visual observation, and reasoning. Yet these stages are typically evaluated on different benchmarks and distributions, making it difficult to determine where hallucinations originate. We first organize existing benchmarks around these stages and show that their scores provide inconsistent diagnostic signals: stronger stage-level performance does not reliably imply lower downstream hallucination, and even benchmarks targeting the same capability can disagree.

We therefore introduce a causal stage-intervention protocol that overwrites individual stages while holding the downstream task fixed. Across 60,008 runs on three video-agent architectures, we find that grounding is the dominant source of downstream error, with roughly four times the causal impact of corrupting visual observations. Successful grounding depends primarily on locating the correct region rather than precise temporal overlap, explaining why standard mIoU metrics poorly predict downstream reliability. We further find that incorrect evidence is substantially more harmful than missing evidence. Finally, auditing existing benchmarks against these interventions reveals that their scores do not reliably predict causal cascade sensitivity and can fail under distribution shift. These results motivate intervention-based, stage-aware evaluation for trustworthy video agents.
\end{abstract}

\section{Introduction}
\label{sec:intro}

\begin{figure}[t]
    \centering
    \includegraphics[width=\columnwidth]{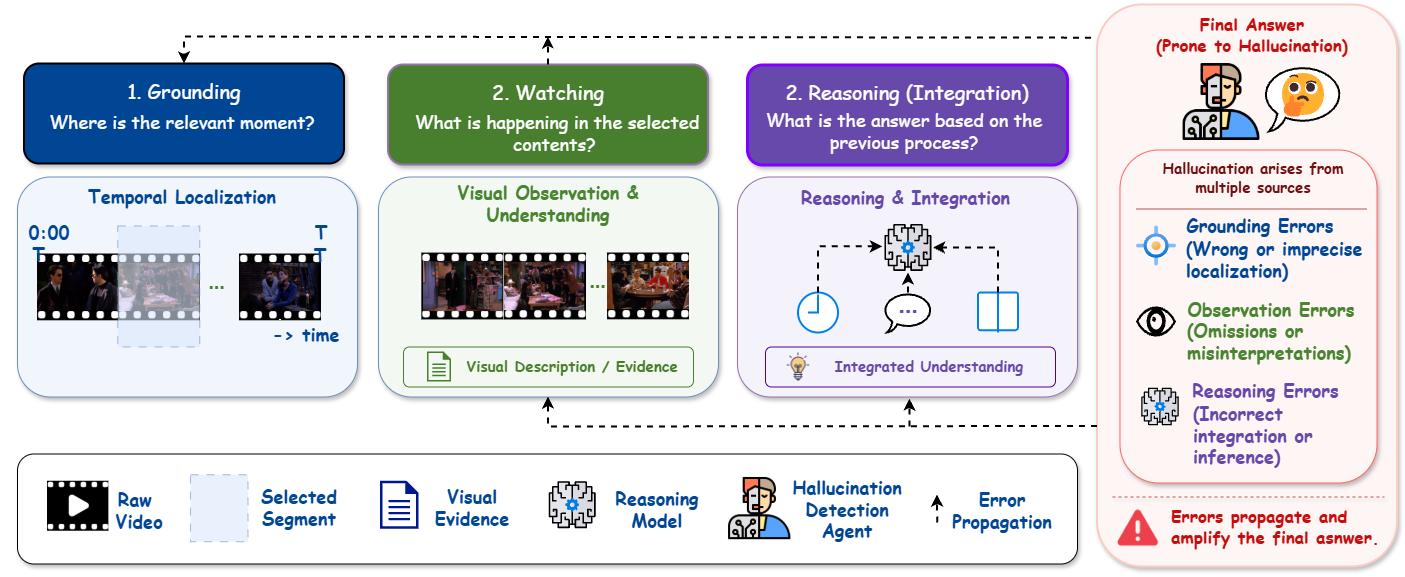}
    \caption{
    A three-stage view of video understanding.
    Grounding localizes relevant moments,
    Watching extracts visual evidence, and
    Reasoning integrates that evidence into an answer.
    Downstream failures may originate at any stage and propagate through the pipeline.
    }
    \label{fig:pipeline}
\end{figure}

Video understanding is increasingly moving from monolithic Video LLMs toward agentic systems that decompose long-video reasoning into specialized stages~\cite{tang2023videounderstanding,maaz2024videochatgpt,zhang2023videollama,lin2024videollava,li2024llamavid,wang2024videoagentlong,liu2025videomind,zhang2026eva,liu2025longvideoagent}. A typical video agent first \textbf{grounds} the question to relevant moments, then \textbf{watches} those moments to extract visual evidence, and finally reasons over the evidence to produce an answer (Figure~\ref{fig:pipeline}). This decomposition improves flexibility and scalability, but also creates multiple points of failure: an agent may retrieve the wrong moment, misinterpret the right one, or reason incorrectly over otherwise valid evidence. Errors can therefore propagate across stages before appearing in the final answer.

Evaluation, however, has not evolved with this pipeline. Existing hallucination benchmarks largely score final outputs~\cite{huang2026videohallucsurvey,elvhalluc2025,videohallucer2024,argus2025}, whereas capability benchmarks separately evaluate grounding~\cite{cgbench2025,nextgqa}, temporal understanding~\cite{lvbench}, or related intermediate behaviors. Because these benchmarks use different datasets, distributions, and protocols, their scores are not directly comparable across stages. A high grounding score, for example, does not establish that grounding errors are unimportant downstream, nor does a poor hallucination score identify which stage caused the failure. 

This raises a fundamental question: \emph{Can existing benchmark scores actually diagnose failures in multi-stage video-agent pipelines?}

We first examine this question observationally. We organize five established benchmarks around Grounding, Watching, and Reasoning and evaluate representative open-weight video agents under controlled inference settings. The resulting profiles are strikingly inconsistent: stronger upstream scores do not reliably imply better downstream hallucination behavior, benchmarks assigned to the same stage can move in opposite directions under the same inference change, and hallucination metrics can produce anomalous values on agentic architectures. These inconsistencies expose a more fundamental limitation: when stages are evaluated on different data, cross-stage comparisons remain correlational and cannot determine which stage actually changes the final outcome.

We therefore replace cross-benchmark correlation with a \textbf{controlled stage-intervention protocol}. Using 2,522 questions over 227 long videos with ground-truth temporal evidence, we overwrite one stage while keeping the downstream task fixed. For Grounding, we substitute oracle, random, off-target, and progressively dilated temporal windows; for Watching, we systematically omit or replace frames within the correct window. Across 60,008 intervened runs over three architecturally distinct agents, this design directly measures how changing a stage changes downstream performance. We further complement answer accuracy with an evidence-level diagnosis that tests whether apparently correct answers are actually supported by the video.

The interventions reveal a consistent causal picture. Grounding dominates downstream error: replacing a length-matched random window with the oracle evidence span improves accuracy by 17--18 points across all three architectures, roughly four times the effect of substantial frame corruption during Watching. Yet deployed grounding modules recover only about one quarter of this available gain. Moreover, being on-target matters far more than being temporally precise: expanding an oracle span by up to $8\times$ incurs essentially no loss, explaining why overlap-based metrics such as mIoU can poorly reflect downstream utility. Finally, wrong evidence is more harmful than missing evidence: agents tolerate severe frame omission but degrade when plausible frames from elsewhere in the video replace the correct evidence. Consistent with these results, our audit finds no existing benchmark score that detectably predicts causally measured cascade sensitivity. Evidence-level diagnosis further reveals that 42--65\% of correct answers can be unsupported by the video, while a hallucination metric that discriminates on its native distribution can silently saturate under domain shift.

Our contributions are threefold:
\begin{enumerate}
    \item We provide a stage-aligned analysis of existing video evaluation, showing that fragmented benchmark scores cannot reliably diagnose where failures arise in multi-stage video agents.

    \item We introduce a causal stage-intervention protocol that overwrites individual pipeline stages on a shared item set while holding the downstream task fixed, enabling direct measurement of stage-level effects across agent architectures.

    \item Using this causal reference, we reveal systematic gaps in current evaluation: grounding dominates downstream error, while deployed grounders capture only a small fraction of the available headroom; on-target evidence matters more than precise temporal localization; incorrect evidence is more damaging than missing evidence; and existing answer- and hallucination-level scores can fail to reflect causal reliability.
\end{enumerate}
\section{Preliminary Study}
\label{sec:setup}

\subsection{Pipeline Formalization and Parameters}
\label{sec:pipeline}

We abstract modern video agents into three functional stages:
\textbf{Grounding} ($\stepG$), which localizes question-relevant temporal evidence;
\textbf{Watching} ($\stepW$), which extracts visual information from the selected content; and
\textbf{Reasoning} ($\stepR$), which integrates this information with the question to produce the final answer.
These stages need not correspond to explicit modules---different agents may implement them through retrieval, frame selection, visual descriptions, latent representations, or tool use---but they provide a common interface for stage-aligned evaluation.
The pipeline is sequential, $\stepG \rightarrow \stepW \rightarrow \stepR$, so upstream errors may affect downstream predictions.

For the preliminary study, we vary two inference-time controls shared by the evaluated systems:
the number of sampled frames $\settingF$ and the maximum tool-call budget $\settingT$.
$\settingF$ controls the amount of visual evidence exposed to the model; we use
$\settingF\in\{16,32,64\}$ for Video-R1, VideoMind, and EVA.
$\settingT$ limits the number of agentic tool invocations and is varied from 4 to 16 for EVA; the other models do not expose an equivalent control.
We focus on these parameters because they can be changed at inference time without modifying model weights, enabling controlled comparisons across configurations.

\subsection{Models, Benchmarks and Metrics}
\label{sec:models}

\begin{wraptable}{r}{0.64\columnwidth}
\vspace{-12pt}
\centering
\scriptsize
\setlength{\tabcolsep}{2.5pt}
\renewcommand{\arraystretch}{0.95}
\begin{tabular}{c c c c}
\toprule
\textbf{Stage} & \textbf{Benchmark} & \textbf{Scale} & \textbf{Metric} \\
\midrule
\multirow{2}{*}{Grounding}
& \nextgqa{}~\citep{nextgqa2024} & 990 / 5,553 & Acc, mIoU \\
& \cgbench{}~\citep{cgbench2025} & 1,219 / 12,129 & Acc, mIoU \\
\midrule
\multirow{2}{*}{Watching}
& \lvbench{} (T.G.)~\citep{lvbench} & 72 / 179 & Acc \\
& \argus{}~\citep{argus2025} & 500 / 500 & $\ArgusCostH{},\ArgusCostO{}$ \\
\midrule
\multirow{3}{*}{Reasoning}
& \argus{}~\citep{argus2025} & 500 / 500 & $\ArgusCostH{},\ArgusCostO{}$ \\
& \elvhalluc{}~\citep{elvhalluc2025} & 200 / 3,600 & Acc, Diff, SAH \\
& VideoHallucer~\citep{videohallucer2024} & 948 / 1,800 & Basic, Halluc, Both \\
\bottomrule
\end{tabular}

\caption{Stage-aligned benchmarks and metrics.}
\label{tab:benchmarks}
\vspace{-6pt}
\end{wraptable}

We evaluate three representative open-weight systems spanning distinct video-agent designs:
\videor{}~\citep{videor12025}, a monolithic Qwen2.5-VL-based reasoner;
\videomind{}~\citep{liu2025videomind}, which separates stages with role-specific LoRA modules; and
\eva{}~\citep{zhang2026eva}, a multi-agent framework with explicit localization, observation, and answering components.
All experiments use publicly released checkpoints without additional fine-tuning.
For \videomind{} and \eva{}, intermediate stage outputs are directly exposed; for \videor{}, we obtain them through structured prompting.
The same systems are used in the causal intervention study in Section~\ref{sec:causal}.

We organize existing benchmarks by the pipeline stage they primarily probe, as summarized in Table~\ref{tab:benchmarks}.
Grounding is evaluated with \nextgqa{} and \cgbench{}, Watching with the temporal-grounding subset of \lvbench{} and \argus{}, and downstream hallucination/reasoning with \argus{}, \elvhalluc{}, and VideoHallucer.
\argus{} spans Watching and Reasoning because it evaluates the faithfulness of generated visual descriptions.


\subsection{Results and Limits of Fragmented Evaluation}
\label{sec:limitations}

We first examine whether existing stage-aligned benchmarks provide consistent
diagnostic signals for hallucination in video agents.

Table~\ref{tab:preliminary-run} shows that capability scores and hallucination
scores do not align consistently across stages.
Strong Grounding or Watching performance does not imply lower downstream
hallucination: for example, \eva{} achieves the best CG-Bench accuracy and
LVBench score, yet performs poorly on ELV-Halluc and the hallucination subset
of VideoHallucer.
Conversely, models with weaker standard capability scores can be substantially
more robust on hallucination-oriented evaluation.
ARGUS provides a useful intermediate signal by directly evaluating hallucination
and omission in visual descriptions, but it still operates on a separate dataset.

\begin{table*}[tbh]
\centering
\scriptsize
\setlength{\tabcolsep}{1.8pt}
\renewcommand{\arraystretch}{0.92}

\resizebox{\textwidth}{!}{
\begin{tabular}{lccccc cc ccc ccc}
\toprule
\multirow{3}{*}{\textbf{Method}} &
\multicolumn{4}{c}{\textbf{Grounding}} &
\multicolumn{1}{c}{\textbf{Watching}} &
\multicolumn{2}{c}{\textbf{W $\leftrightarrow$ H}} &
\multicolumn{6}{c}{\textbf{Hallucination}} \\
\cmidrule(lr){2-5}
\cmidrule(lr){6-6}
\cmidrule(lr){7-8}
\cmidrule(lr){9-14}
&
\multicolumn{2}{c}{\textbf{CG-Bench}} &
\multicolumn{2}{c}{\textbf{NeXT-GQA}} &
\textbf{LVBench*} &
\multicolumn{2}{c}{\textbf{ARGUS}} &
\multicolumn{3}{c}{\textbf{ELV-Halluc}} &
\multicolumn{3}{c}{\textbf{VideoHallucer}} \\
\cmidrule(lr){2-3}
\cmidrule(lr){4-5}
\cmidrule(lr){6-6}
\cmidrule(lr){7-8}
\cmidrule(lr){9-11}
\cmidrule(lr){12-14}
&
\textbf{Acc$\uparrow$} & \textbf{mIoU$\uparrow$} &
\textbf{Acc$\uparrow$} & \textbf{mIoU$\uparrow$} &
\textbf{Acc$\uparrow$} &
\textbf{$\ArgusCostH{}\downarrow$} & \textbf{$\ArgusCostO{}\downarrow$} &
\textbf{Acc$\uparrow$} & \textbf{Diff$\downarrow$} & \textbf{SAH$\downarrow$} &
\textbf{Basic$\uparrow$} & \textbf{Halluc$\uparrow$} & \textbf{Both$\uparrow$} \\
\midrule
Qwen2-VL-7B   & 31.37 & -    & 74.59 & -     & 54.79 & 55.52 & 82.09 & 8.1  & 5.8  & 6.1  & 76.42 & 64.25 & 46.58 \\
Qwen2.5-VL-7B & 32.24 & -    & 72.98 & -     & 58.10 & 51.19 & 81.57 & 14.2 & 4.5  & 5.1  & 74.58 & 68.83 & 49.58 \\
Video-R1       & 35.32 & 1.44 & 75.38 & 21.14 & 59.82 & 66.15 & 83.58 & 15.2 & -0.3 & -0.3 & 81.08 & 55.50 & 42.58 \\
VideoMind-7B   & 38.40 & 7.10 & 76.79 & 31.40 & 54.79 & 55.63 & 83.06 & 11.4 & 4.2  & 4.6  & 76.25 & 64.58 & 46.83 \\
EVA            & 40.20 & 4.51 & 68.99 & 26.87 & 60.27 & 53.10 & 82.81 & 0.6  & -0.5 & -0.5 & 88.25 & 45.83 & 39.08 \\
\midrule
\textbf{Avg.}  & 35.51 & 4.35 & 73.75 & 26.47 & 57.55 & 56.32 & 82.62 & 9.90 & 2.74 & 3.00 & 79.32 & 59.80 & 44.93 \\
\bottomrule
\end{tabular}
}

\caption[Stage-aligned benchmark results under default inference settings.]{
Stage-aligned benchmark results under default inference settings.\protect\footnotemark
}
\label{tab:preliminary-run}
\end{table*}

\footnotetext{
Default settings are Video-R1 ($f{=}16$), VideoMind-7B ($f{=}32$), and EVA ($\tau{=}4,f{=}32$).
LVBench* denotes the Temporal Grounding subset; ARGUS reports hallucination/omission costs; VideoHallucer reports Basic/Halluc/Both accuracy.
}


Overall, the benchmark profile can rank individual capabilities, but it does not
reveal where a downstream hallucination originates. Detailed per-model comparisons are provided in Appendix~\ref{app:preliminary}. Additional parameter-sweep results in Appendix~\ref{app:preliminary} further show that changing the sampled-frame count or tool-call budget produces benchmark- and model-dependent effects, with no consistent improvement in downstream hallucination robustness.

\textbf{Why Fragmented Evaluation Is Not Enough?} The preliminary study exposes two problems. First, stage-level capability scores are only weakly connected to downstream
hallucination.
Second, conclusions do not reliably transfer across datasets, even within the
same nominal stage.
The anomalous near-zero or negative ELV-Halluc readings for \eva{} and
\videor{} make this ambiguity particularly clear: from the score alone, we
cannot tell whether the model failed or the metric did.

More fundamentally, all of these comparisons are correlational and are measured
on different datasets.
They can reveal disagreement, but cannot determine which pipeline stage caused
a downstream error or how much correcting that stage would help.
Section~\ref{sec:causal} therefore replaces cross-benchmark comparison with
controlled stage interventions on a single item set.
\section{Causal Stage Intervention}
\label{sec:causal}
The preliminary study reveals disagreement between stage-level and hallucination
scores, but cannot identify its source because each stage is evaluated on a
different dataset. We therefore move to a controlled design in which the
\emph{same items and downstream task} are retained while one upstream stage is
overwritten with a known value. This allows us to directly measure how errors at
each stage affect the final prediction. Overall, we conduct \textbf{60{,}008
intervened agent runs} and \textbf{5{,}784 evidence-level diagnoses} across
three architectures.

\subsection{Intervention Protocol}
\label{sec:protocol-causal}

\textbf{Data and grounding interventions.}
We construct the intervention set from \cgbench{}~\citep{cgbench2025}, which
provides both ground-truth temporal evidence spans (\texttt{clue\_intervals})
and a native \emph{Hallucination} question category. We use all \textbf{2{,}522
questions} from the 227 videos containing at least one Hallucination item:
444 Hallucination questions and 2{,}078 questions from eleven other categories.
Because both groups concern the same videos, comparisons are not confounded by
video distribution. The median video lasts 26 minutes, whereas the median clue
span is only 10\,s (0.64\% of the video).
For each item, we replace the agent's temporal window while leaving the remaining
pipeline unchanged. The conditions are \condfull{} (whole video),
\condpred{} (the agent's predicted span), \condoracle{} (ground-truth clue
span), \condrandom{} (a length-matched random span), and \condoff{} (a
length-matched span in the largest evidence-free gap). We additionally dilate
the oracle span by $2\times/4\times/8\times/16\times$ on 300 items to test how
precise grounding must be.

\textbf{Watching interventions and implementation.}
With grounding fixed to \condoracle{}, we corrupt the sampled frames within the
correct window. \emph{Omission} removes a fraction $\rho$ of frames, whereas
\emph{fabrication} replaces them with frames sampled at least 60\,s away from
the same video while preserving their original timestamp labels, with
$\rho\in\{0.25,0.5,0.75\}$. Both interventions operate on the sampled frame
list shared by all three systems.
The intervention is inserted at the corresponding interface of each
architecture: \eva{}'s requested windows are restricted to the assigned span,
\videomind{}'s answerer receives that span instead of the grounder's prediction,
and \videor{}, which has no explicit grounding module, receives it directly
through its frame list and serves as the monolithic control. Thus,
\condpred{} is undefined for \videor{}.

\textbf{Statistical protocol.}
We report 95\% paired bootstrap intervals clustered by video (227 clusters).
Paired binary outcomes are tested with exact McNemar tests and Holm correction
within each contrast family; unparseable outputs count as incorrect and are
reported separately as abstentions. As a pre-specified sanity check,
\condoracle{} must outperform \condrandom{} for every agent, while
\condoff{} must not outperform \condrandom{}. Both conditions hold
(Table~\ref{tab:causal-conditions}), confirming that the interventions are
effectively propagated to the models.

\begin{table}[tbh]
\centering
\small
\setlength{\tabcolsep}{4.5pt}
\begin{tabular}{lccc}
\toprule
\textbf{Grounding condition} & \eva{} & \videomind{} & \videor{} \\
\midrule
\condoracle{} (GT clue span) & \textbf{46.6} {\tiny[44.2, 48.9]} & \textbf{49.3} {\tiny[47.0, 51.6]} & \textbf{47.2} {\tiny[44.9, 49.6]} \\
\condpred{} (agent as deployed) & 34.1 {\tiny[31.8, 36.4]} & 35.9 {\tiny[33.6, 38.2]} & --- \\
\condfull{} (whole video) & 27.0 {\tiny[25.1, 28.9]} & 34.8 {\tiny[32.5, 37.1]} & 33.3 {\tiny[31.1, 35.4]} \\
\condrandom{} (length-matched) & 29.0 {\tiny[27.0, 31.1]} & 31.4 {\tiny[29.3, 33.4]} & 30.6 {\tiny[28.7, 32.6]} \\
\condoff{} (length-matched) & 27.2 {\tiny[25.3, 29.1]} & 30.6 {\tiny[28.4, 33.0]} & 30.7 {\tiny[28.6, 32.8]} \\
\midrule
\multicolumn{4}{l}{\textit{Headroom decomposition (paired contrasts, pp)}} \\
\condoracle{} $-$ \condrandom{} \hfill (total headroom) & $+$17.6 {\tiny[14.9, 20.2]} & $+$17.9 {\tiny[15.8, 20.0]} & $+$16.6 {\tiny[14.5, 18.8]} \\
\condpred{} $-$ \condrandom{} \hfill (captured) & $+$5.0 {\tiny[2.7, 7.3]} & $+$4.5 {\tiny[2.6, 6.4]} & --- \\
\condoracle{} $-$ \condpred{} \hfill (left on the table) & $+$12.5 {\tiny[10.2, 14.8]} & $+$13.4 {\tiny[11.3, 15.5]} & --- \\
\bottomrule
\end{tabular}

\vspace{4pt}
\caption{Causal grounding intervention on 2{,}522 paired \cgbench{}
items from 227 video clusters. Top: accuracy (\%) with clustered 95\% bootstrap
CIs. Bottom: paired contrasts. All \condoracle{}$-$\condrandom{} differences
are significant at $p<0.001$. The deployed
grounders capture only $\sim$25\% of the available oracle headroom.}
\label{tab:causal-conditions}
\end{table}

\begin{figure}[t]
    \centering

    \begin{subfigure}[t]{0.63\textwidth}
        \centering
        \includegraphics[width=\linewidth]{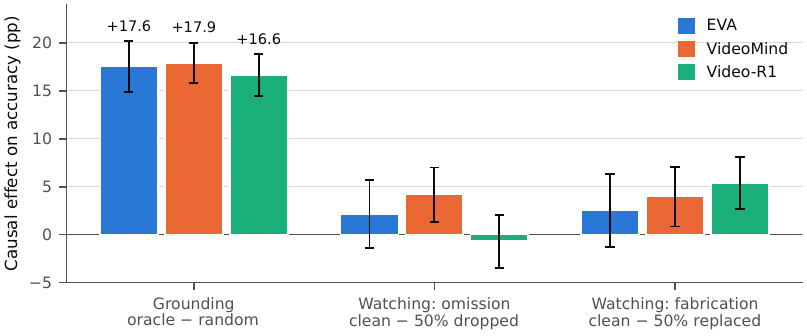}
        \caption{Causal effects of grounding and watching interventions.}
        \label{fig:stage_sensitivity}
    \end{subfigure}
    \hfill
    \begin{subfigure}[t]{0.36\textwidth}
        \centering
        \includegraphics[width=\linewidth]{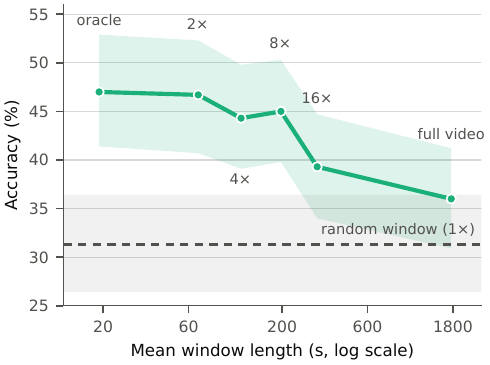}
        \caption{Oracle-window dilation accuracy. }
        \label{fig:dose_response}
    \end{subfigure}

    \caption{Grounding determines where the model looks, but does not
    need to be temporally precise.
    \textbf{Left:} replacing a length-matched random window with the oracle span
    improves accuracy by 17--18\,pp across all three architectures, substantially
    more than corrupting frames within the correct window.
    \textbf{Right:} expanding the oracle window up to $8\times$ causes little
    degradation; performance drops only once the window becomes very broad.}
    \label{fig:causal_grounding}
\end{figure}

\subsection{Analysis and Findings of Causal Stage Intervention}
\textbf{Grounding Dominates Downstream Error. }
Table~\ref{tab:causal-conditions} and
Figure~\ref{fig:causal_grounding} show a clear asymmetry between grounding and
watching. Replacing a length-matched random window with the oracle span improves
accuracy by 16.6--17.9\,pp across all three architectures ($p<0.001$). By
contrast, dropping or replacing half of the frames \emph{within the correct
window} changes accuracy by at most about 5\,pp. Thus, for long-video
understanding, \emph{where} the model looks matters substantially more than
moderate corruption of \emph{what} it sees once the relevant moment has been
found.

Current grounding modules exploit only a small fraction of this opportunity.
Relative to a random window, the oracle span provides roughly 18\,pp of
headroom, whereas the deployed grounders recover only 5.0\,pp for \eva{} and
4.5\,pp for \videomind{}. The remaining gap is 12.5 and 13.4\,pp,
respectively. In particular, \videomind{}'s predicted grounding performs only
marginally better than showing the whole video, while \eva{} obtains a clearer
gain. Across these different architectures, the same pattern emerges:
grounding is both the most consequential stage and the stage with the largest
remaining room for improvement.

\textbf{Wrong evidence is more harmful than missing evidence.}
The watching interventions further show that agents tolerate missing visual
evidence better than misleading evidence. Even after dropping 75\% of the
frames within the correct window, the degradation is small and not consistently
significant. Replacing those frames with plausible content from elsewhere in
the same video causes a clearer drop in accuracy (Appendix~\ref{app:ladders}).
This suggests that grounding errors are harmful not simply because they remove
useful frames, but because they can supply convincing evidence from the wrong
moment.

\textbf{Why the unified design matters.}
These effects are difficult to recover from cross-benchmark comparisons.
On the shorter \drvbench{} slice, for example, predicted grounding does not
outperform simply exposing the whole video for either \eva{} or \videomind{}.
This is consistent with the dilation result: when the video itself is relatively
short, a broad but on-target context can be sufficient. More generally,
comparing grounding and hallucination scores across different datasets cannot
separate such distribution effects from genuine stage sensitivity. The
intervention design avoids this confound by changing one stage while evaluating
the same downstream items.
\section{Auditing the Benchmark Ecosystem Against the Causal Standard}
\label{sec:audit}

The intervention protocol provides more than stage-level error attribution. For
each item and system configuration, it gives a direct measure of how much a
particular stage affects the final prediction. This allows us to revisit the
observation in Section~\ref{sec:limitations} that existing benchmark scores are
only weakly connected across stages, and ask a more concrete question:
\emph{do existing benchmark scores predict how sensitive the final answer is
to a causal intervention on that stage?}

\paragraph{Cascade sensitivity.}
For each item, we define cascade sensitivity as the change in answer accuracy
when grounding is replaced by the oracle span rather than a
length-matched random span. At the system level, we average this quantity over
items. Because sensitivity is measured on the same 2{,}522 items used in
Section~\ref{sec:causal}, it serves as a causal reference quantity rather than
another benchmark score.

\begin{figure}[tbh]
    \centering
    \includegraphics[width=0.95\textwidth]{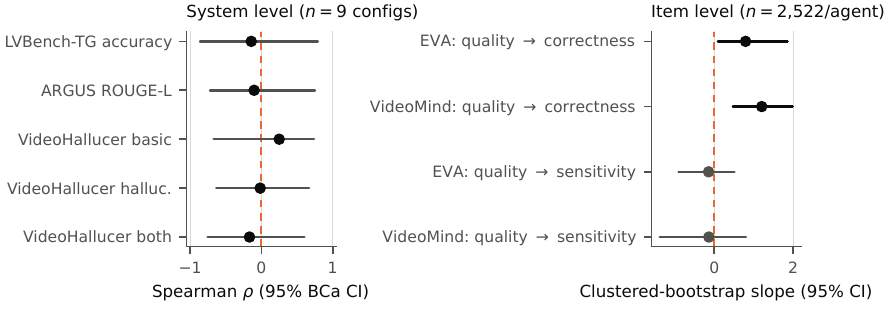}
    \caption{\textbf{No existing benchmark score detectably predicts causally
    measured cascade sensitivity.} Left: Spearman correlation between five
    retained stage-aligned benchmark scores (recomputed from per-item outputs
    of Section~\ref{sec:limitations}) and causal sensitivity across nine
    agent$\times$frame-budget configurations; every 95\% BCa interval spans
    zero. Right: within-agent, per-item clustered regression. Grounding quality
    (frame-level evidence rate) predicts \emph{whether the agent answers
    correctly} (top two rows, intervals exclude zero), but does not predict
    \emph{whether grounding causally determines the answer} (bottom two rows).
    The same pattern appears for both agents with an explicit grounding stage.}
    \label{fig:audit_forest}
\end{figure}

\subsection{No benchmark score predicts what causally matters}
\label{sec:audit-results}

Figure~\ref{fig:audit_forest} shows the same pattern at both the item and
system levels: conventional benchmark scores track \emph{performance}, but not
\emph{causal dependence}.

\textbf{Item level.}
Better grounding is associated with higher final-answer accuracy for both
\eva{} and \videomind{}, confirming the familiar leaderboard signal that agents
tend to answer better when they localize more relevant evidence. However,
grounding quality does not predict cascade sensitivity: items with better
grounding are no more likely to be those whose answers actually change when
grounding is corrected. The regression slopes for correctness are clearly
positive, whereas those for causal sensitivity remain near zero
(Figure~\ref{fig:audit_forest}). In other words, standard grounding metrics can
tell us whether an agent is doing well, but not whether grounding is the stage
responsible for that outcome.

\textbf{System level.}
The same conclusion holds across agent and frame-budget configurations. None of
the retained stage-aligned benchmarks---\lvbench{}-TG, \argus{}, or the three
VideoHallucer scores---shows a detectable association with causally measured
cascade sensitivity. Despite measuring grounding, visual faithfulness, and
hallucination from different perspectives, these scores do not identify which
systems are more sensitive to grounding errors.

Taken together, the two analyses expose a fundamental gap between benchmark
performance and diagnostic value: \textbf{existing scores can rank systems by
how well they perform, but they do not reveal how strongly a particular stage
causally determines the final answer.}

\subsection{Beyond right and wrong: correct answers can still be hallucinated}
\label{sec:drv-readout}

Final-answer accuracy cannot distinguish a correct answer supported by the video from one reached without sufficient visual evidence. To expose this difference, we apply \drv{}, a stage-aware
hallucination diagnosis framework that verifies claims in an agent's answer
against sampled visual evidence and attributes failures to perceptive,
temporal, or cognitive causes.\footnote{Reference withheld for double-blind
review.} We run the diagnosis on 900 outputs from the \drvbench{} long-video
slice and 4{,}884 outputs from \cgbench{}.

\begin{figure}[tbh]
    \centering
    \includegraphics[width=0.95\textwidth]{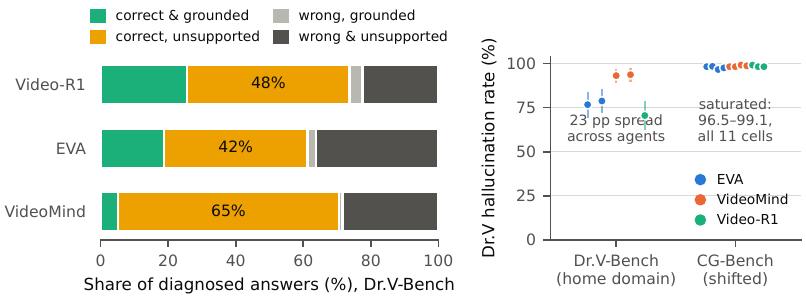}
    \caption{\textbf{Left:} evidence-level diagnosis on \drvbench{}. A large
    fraction of answers are \emph{correct but unsupported}: 42--65\% of correct
    answers lack sufficient visual evidence, a failure mode hidden by
    final-answer accuracy.
    \textbf{Right:} \drv{} clearly separates agents on its home domain, but
    saturates on \cgbench{} across agents and intervention conditions,
    indicating that the same hallucination metric can lose discriminative
    power under distribution shift.}
    \label{fig:drv_readout}
\end{figure}

Figure~\ref{fig:drv_readout} reveals a substantial gap between answer
correctness and evidential support. Across the three agents, \textbf{42--65\%
of correct answers are not sufficiently supported by the video}. These answers
would be counted as successful under a conventional accuracy metric, even
though the model did not establish the visual evidence needed to justify them.

This difference is important for reliability evaluation. Agents with similar
final-answer accuracy can differ substantially in how much of that accuracy is
actually grounded in the video. In our results, \drv{} separates the agents by
more than 20 percentage points in hallucination rate despite much smaller
differences in answer accuracy. Thus, final-answer correctness and evidential
support capture different aspects of model behavior.

\textbf{A correct answer is therefore not necessarily a reliable answer.}
Evaluation should distinguish whether the model arrives at the right prediction
from supported visual evidence, rather than treating all correct outputs as
equally successful.

\subsection{Metric failure versus model failure}
\label{sec:metric-failure}

The preliminary study shows that anomalous benchmark scores do not necessarily
imply model failure. On \elvhalluc{}, \eva{} and \videor{} obtain near-zero or
negative scores, far outside the regime observed for the monolithic models on
which the benchmark was validated.

Our intervention results show how such failures can arise. \drv{} clearly
separates agents on its native \drvbench{} distribution, but saturates near
100\% on \cgbench{} across agents and intervention conditions
(Figure~\ref{fig:drv_readout}, right). More importantly, replacing a random
grounding window with the oracle span improves answer accuracy by about
17\,pp while barely changing the \drv{} score. The metric therefore loses
discriminative power under the shifted evaluation regime. The likely cause is a mismatch between the metric and the target setting:
\cgbench{} produces long free-form answers with many checkable claims, while
the diagnosis relies on sparsely sampled frames from long videos, making a
positive hallucination verdict almost inevitable. A similar architectural
mismatch may explain the anomalous \elvhalluc{} scores for tool-calling agents.

An extreme benchmark score can therefore reflect metric failure rather than
model failure.






\subsection{Implications for trustworthy evaluation}
\label{sec:implications}

These results suggest three practical changes to how multi-stage video systems
should be evaluated.

\textbf{(i) Measure whether grounding is on target, not only how tightly it
overlaps the annotation.}
Finding~3 shows that substantial temporal dilation around the correct evidence
has little effect on downstream accuracy, even though it can strongly reduce
overlap-based measures such as mIoU. Grounding metrics should therefore
distinguish between missing the relevant event and selecting a somewhat wider
window around it. One useful criterion is whether the annotated evidence is
contained within the selected window at a practically useful dilation.

\textbf{(ii) Check that a metric remains discriminative in the target regime.}
Before interpreting a hallucination score, the metric should be tested against
a controlled manipulation known to change model behavior. An oracle-span
intervention provides one inexpensive and reusable probe. If a metric remains
nearly constant despite a large and independently verified change in answer
accuracy, as in Section~\ref{sec:metric-failure}, its numerical value should
not be treated as a reliable measure of model behavior in that regime.

\textbf{(iii) Evaluate evidential support in addition to answer correctness.}
The 42--65\% rate of correct-but-unsupported answers shows that correctness
alone is insufficient for assessing reliability. Evaluation should therefore
record not only whether an answer is correct, but also whether the visual
evidence available to the agent actually supports it. Stage-annotated datasets
such as \cgbench{}, which provide clue intervals, make this type of
evidence-level evaluation feasible at benchmark scale.
\section{Conclusion}
\label{sec:conclusion}

Existing video benchmarks can rank capabilities, but they do not reliably diagnose where failures arise or which stages causally determine downstream outcomes in multi-stage video agents. By replacing cross-benchmark comparison with controlled stage interventions, we find that grounding is the dominant downstream bottleneck, that being on-target matters more than precise temporal overlap, and that wrong evidence is substantially more harmful than missing evidence. 
Our benchmark audit further shows that conventional capability- and hallucination-level scores do not reliably predict causal cascade sensitivity.
These results suggest that trustworthy evaluation should move beyond isolated benchmark scores toward intervention-based, evidence-aware evaluation of video agents.

\textbf{Limitations.}
First, the \cgbench{} intervention results measure downstream answer \emph{error} rather than hallucination defined as unsupported content. The latter is examined separately with \drvbench{}, where the diagnosis instrument retains discriminative power. Second, we evaluate one checkpoint per architecture, so cross-agent dissociations (e.g., \videor{}'s sensitivity to fabrication) cannot separate architecture from model identity. Third, the system-level audit covers nine configurations and its wide intervals support only ``no detectable
relationship''; the item-level analysis ($n{=}2{,}522$ per agent) is the powered version of the claim. Fourth, \eva{}'s \condfull{} condition required a forced-answer prompt to suppress tool-denied refusals (flagged per-run; Appendix~\ref{app:protocol-details}), and multi-interval clues (11\% of items) are served as their temporal hull.

\clearpage
\bibliographystyle{plainnat}
\bibliography{main}
\newpage
\section*{NeurIPS Paper Checklist}

\begin{enumerate}

\item {\bf Claims}
    \item[] Question: Do the main claims made in the abstract and introduction accurately reflect the paper's contributions and scope?
    \item[] Answer: \answerYes{} 
    \item[] Justification: Contributions are correctly reflected in abstract and introduction.

\item {\bf Limitations}
    \item[] Question: Does the paper discuss the limitations of the work performed by the authors?
    \item[] Answer: \answerYes{} 
    \item[] Justification: Limitations and corresponding future works are discussed.

\item {\bf Theory assumptions and proofs}
    \item[] Question: For each theoretical result, does the paper provide the full set of assumptions and a complete (and correct) proof?
    \item[] Answer: \answerYes{} 
    \item[] Justification: Proof or analysis of the proposed theoretical result are included in the paper.

    \item {\bf Experimental result reproducibility}
    \item[] Question: Does the paper fully disclose all the information needed to reproduce the main experimental results of the paper to the extent that it affects the main claims and/or conclusions of the paper (regardless of whether the code and data are provided or not)?
    \item[] Answer: \answerYes{} 
    \item[] Justification: Model details are disclosed

\item {\bf Open access to data and code}
    \item[] Question: Does the paper provide open access to the data and code, with sufficient instructions to faithfully reproduce the main experimental results, as described in supplemental material?
    \item[] Answer: \answerNo{} 
    \item[] Justification: Code and instructions will be released upon acceptance

\item {\bf Experimental setting/details}
    \item[] Question: Does the paper specify all the training and test details (e.g., data splits, hyperparameters, how they were chosen, type of optimizer) necessary to understand the results?
    \item[] Answer: \answerYes{} 
    \item[] Justification: All details are provided.

\item {\bf Experiment statistical significance}
    \item[] Question: Does the paper report error bars suitably and correctly defined or other appropriate information about the statistical significance of the experiments?
    \item[] Answer: \answerYes{} 
    \item[] Justification: The paper reports standard metrics including error bars.

\item {\bf Experiments compute resources}
    \item[] Question: For each experiment, does the paper provide sufficient information on the computer resources (type of compute workers, memory, time of execution) needed to reproduce the experiments?
    \item[] Answer: \answerYes{} 
    \item[] Justification: All details are provided
    
\item {\bf Code of ethics}
    \item[] Question: Does the research conducted in the paper conform, in every respect, with the NeurIPS Code of Ethics \url{https://neurips.cc/public/EthicsGuidelines}?
    \item[] Answer: \answerYes{} 
    \item[] Justification: the research conform with the NeurIPS Code of Ethics.

\item {\bf Broader impacts}
    \item[] Question: Does the paper discuss both potential positive societal impacts and negative societal impacts of the work performed?
    \item[] Answer: \answerYes{} 
    \item[] Justification: Discussed in conclusion
    
\item {\bf Safeguards}
    \item[] Question: Does the paper describe safeguards that have been put in place for responsible release of data or models that have a high risk for misuse (e.g., pre-trained language models, image generators, or scraped datasets)?
    \item[] Answer: \answerNA{} 
    \item[] Justification: the paper poses no such risks

\item {\bf Licenses for existing assets}
    \item[] Question: Are the creators or original owners of assets (e.g., code, data, models), used in the paper, properly credited and are the license and terms of use explicitly mentioned and properly respected?
    \item[] Answer: \answerYes{} 
    \item[] Justification: existing assets are properly credited

\item {\bf New assets}
    \item[] Question: Are new assets introduced in the paper well documented and is the documentation provided alongside the assets?
    \item[] Answer: \answerNA{} 
    \item[] Justification: the paper does not release new assets.

\item {\bf Crowdsourcing and research with human subjects}
    \item[] Question: For crowdsourcing experiments and research with human subjects, does the paper include the full text of instructions given to participants and screenshots, if applicable, as well as details about compensation (if any)? 
    \item[] Answer: \answerNA{} 
    \item[] Justification: the paper does not involve crowdsourcing nor research with human subjects.

\item {\bf Institutional review board (IRB) approvals or equivalent for research with human subjects}
    \item[] Question: Does the paper describe potential risks incurred by study participants, whether such risks were disclosed to the subjects, and whether Institutional Review Board (IRB) approvals (or an equivalent approval/review based on the requirements of your country or institution) were obtained?
    \item[] Answer: \answerNA{} 
    \item[] Justification: paper does not involve crowdsourcing nor research with human subjects.

\item {\bf Declaration of LLM usage}
    \item[] Question: Does the paper describe the usage of LLMs if it is an important, original, or non-standard component of the core methods in this research? Note that if the LLM is used only for writing, editing, or formatting purposes and does \emph{not} impact the core methodology, scientific rigor, or originality of the research, declaration is not required.
    \item[] Answer: \answerYes{} 
    \item[] Justification: the core research in this paper involves LLMs as important components.

\end{enumerate}

\appendix
\clearpage
\section{Additional Analysis of the Preliminary Study}
\label{app:preliminary}

This appendix expands the observational results reported in
Section~\ref{sec:limitations}. Table~\ref{tab:preliminary-run} remains in the main
paper; here we provide additional interpretation of the benchmark profiles and
parameter sweeps.

\subsection{Detailed Benchmark Comparisons}

\textbf{Grounding.}
The two Grounding benchmarks distinguish temporal localization capability, but
their rankings do not translate consistently to hallucination robustness.
\videomind{} achieves the strongest grounding mIoU, while \eva{} obtains the
best CG-Bench accuracy. However, \eva{} performs substantially worse on
hallucination-oriented metrics, indicating that stronger measured grounding
does not necessarily correspond to lower downstream hallucination.

\textbf{Watching.}
A similar disconnect appears for visual observation.
\eva{} obtains the highest LVBench accuracy but weak hallucination scores,
whereas Qwen2.5-VL-7B performs slightly worse on LVBench yet much better on
ELV-Halluc and VideoHallucer.
ARGUS is more directly related to hallucination because it measures fabricated
and omitted content in generated visual descriptions, but it still provides only
an observational signal on its own distribution.

\textbf{Hallucination metrics.}
ELV-Halluc and VideoHallucer show some common model-level tendencies, but their
absolute behavior differs substantially.
In particular, ELV-Halluc returns near-zero or negative values for \eva{} and
\videor{}, well outside its usual operating regime.
These values are therefore difficult to interpret without an independent
reference for whether the underlying model behavior actually changed.

\subsection{Detailed Parameter-Sweep Analysis}

\textbf{Frame count.}
For \eva{}, increasing the number of sampled frames raises CG-Bench long-video
accuracy from 36.93\% to 41.00\%, but decreases NeXT-GQA mIoU.
Thus, two Grounding benchmarks react in opposite directions to the same change.
ARGUS varies only mildly for \eva{} and \videomind{}, while VideoHallucer is
largely flat for \eva{}, changes slightly for \videomind{}, and degrades more
substantially for \videor{}.

\textbf{Tool-call budget.}
Increasing \eva{}'s tool-call budget likewise produces no monotonic benefit.
CG-Bench accuracy peaks at an intermediate budget, NeXT-GQA changes only
marginally, and hallucination-oriented metrics show small or inconsistent
responses.
Hence, additional reasoning steps do not reliably improve hallucination
robustness.

\subsection{Detailed Parameter-Sweep Analysis}

\textbf{Frame count.}
For \eva{}, increasing the number of sampled frames raises CG-Bench long-video
accuracy from 36.93\% to 41.00\%, but decreases NeXT-GQA mIoU.
Thus, two Grounding benchmarks react in opposite directions to the same change.
ARGUS varies only mildly for \eva{} and \videomind{}, while VideoHallucer is
largely flat for \eva{}, changes slightly for \videomind{}, and degrades more
substantially for \videor{}.

\textbf{Tool-call budget.}
Increasing \eva{}'s tool-call budget likewise produces no monotonic benefit.
CG-Bench accuracy peaks at an intermediate budget, NeXT-GQA changes only
marginally, and hallucination-oriented metrics show small or inconsistent
responses.
Hence, additional reasoning steps do not reliably improve hallucination
robustness.

\subsection{Inference-Parameter Sweeps}
\label{app:param-sweep}
\begin{figure*}[tbh]
    \centering
    \includegraphics[width=0.85\textwidth]{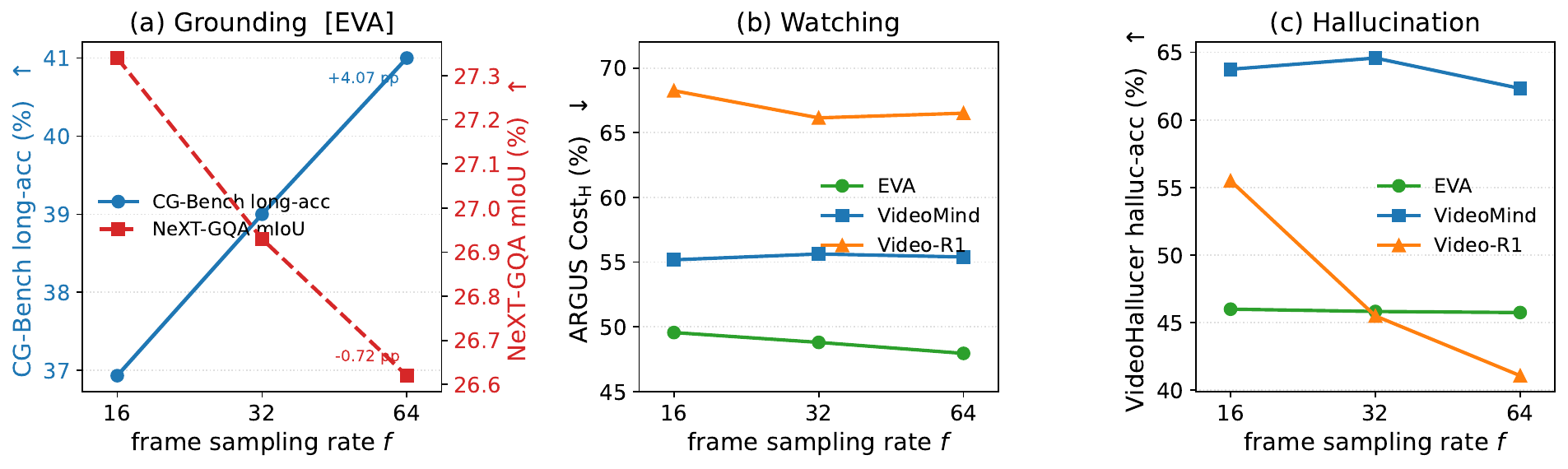}
    \caption{
    Effect of frame sampling rate on Grounding, Watching, and Hallucination benchmarks.
    The same increase in frame count can improve one benchmark while degrading another, revealing benchmark-dependent responses to identical inference changes.
    }
    \label{fig:frame_sweep}
\end{figure*}

\begin{figure}[tbh]
    \centering
    \includegraphics[width=0.49\textwidth]{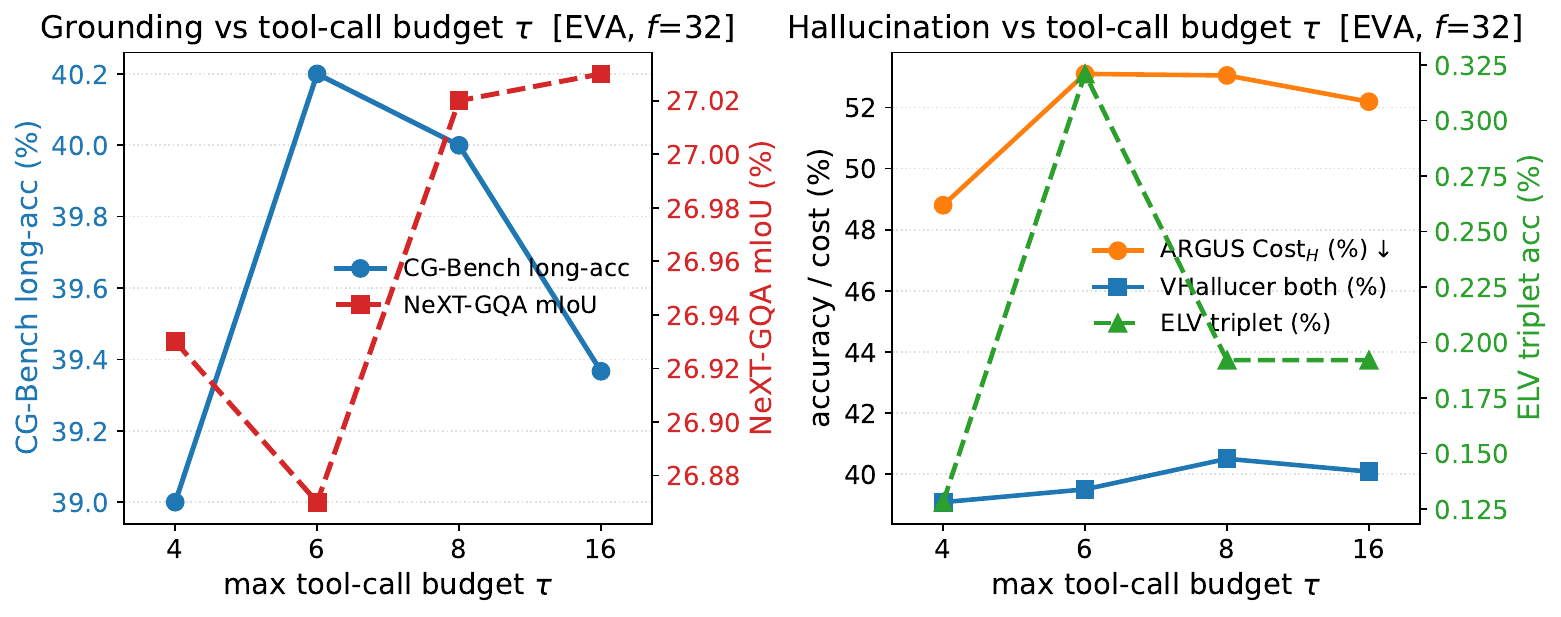}
    \caption{
    Effect of maximum tool-call budget $\tau$ for \eva{} at fixed frame count $\settingF=32$.
    Larger tool budgets do not yield monotonic gains in either Grounding or hallucination metrics.
    }
    \label{fig:tool_sweep}
\end{figure}

Figure~\ref{fig:frame_sweep} shows that the same inference change can lead to
different conclusions across benchmarks.
For \eva{}, increasing the sampled-frame count improves CG-Bench long-video
accuracy but decreases NeXT-GQA mIoU, even though both evaluate Grounding.
Watching and hallucination metrics are similarly model-dependent.

Figure~\ref{fig:tool_sweep} shows the same instability for agentic reasoning:
increasing \eva{}'s tool-call budget does not monotonically improve either
Grounding or hallucination metrics.
More frames or more tool calls therefore cannot be interpreted as uniformly
improving hallucination robustness.

\subsection{Interpretation}

These observations demonstrate instability rather than attribution.
Because each benchmark uses a different dataset and protocol, cross-benchmark
differences conflate model capability with dataset effects.
They therefore motivate, but cannot replace, the controlled intervention study
in Section~\ref{sec:causal}.

\section{Full Intervention Ladders}
\label{app:ladders}

Tables~\ref{tab:app-corruption} and~\ref{tab:app-dilation} report the complete
watching-corruption and dilation ladders. All intervals are 95\% bootstrap CIs
clustered by video. At the strongest corruption rate ($\rho{=}0.75$), the
paired clean$-$corrupted contrasts are: omission $+5.7$\,pp [$-0.3, +11.6$]
(\eva{}) and $+1.7$\,pp [$-3.0, +6.2$] (\videor{}); fabrication $+6.7$\,pp
[$+1.0, +12.6$] (\eva{}) and $+4.3$\,pp [$0.0, +8.9$] (\videor{}). Even
discarding three quarters of the evidence frames inside a correct window is
not reliably detectable, whereas replacing them with plausible frames from
elsewhere in the same video is.

\begin{table}[tbh]
\centering
\small
\setlength{\tabcolsep}{5pt}
\begin{tabular}{llccc}
\toprule
\textbf{Corruption} & $\rho$ & \eva{} & \videomind{} & \videor{} \\
\midrule
none (clean, oracle window) & 0 & 43.2 {\tiny[39.3, 47.0]} & 49.5 {\tiny[45.5, 53.5]} & 46.5 {\tiny[42.6, 50.4]} \\
\midrule
\multirow{3}{*}{omission (frames dropped)}
 & 0.25 & 46.7 {\tiny[41.1, 52.2]} & 50.0 {\tiny[44.3, 55.6]} & 48.0 {\tiny[42.2, 53.9]} \\
 & 0.50 & 41.0 {\tiny[37.2, 44.8]} & 45.3 {\tiny[41.4, 49.3]} & 47.2 {\tiny[43.3, 50.9]} \\
 & 0.75 & 40.3 {\tiny[35.3, 45.4]} & 45.7 {\tiny[40.0, 51.3]} & 43.7 {\tiny[38.1, 49.3]} \\
\midrule
\multirow{3}{*}{fabrication (frames replaced)}
 & 0.25 & 42.7 {\tiny[37.2, 48.4]} & 50.3 {\tiny[44.9, 55.7]} & 45.0 {\tiny[39.7, 50.6]} \\
 & 0.50 & 40.7 {\tiny[36.9, 44.5]} & 45.5 {\tiny[41.4, 49.5]} & 41.2 {\tiny[37.3, 45.0]} \\
 & 0.75 & 39.3 {\tiny[34.5, 44.3]} & 43.0 {\tiny[37.6, 48.5]} & 41.0 {\tiny[35.5, 46.6]} \\
\bottomrule
\end{tabular}
\vspace{4pt}
\caption{Watching-corruption ladder: accuracy (\%) with grounding pinned to the
oracle span and a fraction $\rho$ of in-window frames corrupted
($n{=}600$ at $\rho \in \{0, 0.5\}$, $n{=}300$ otherwise). Fabricated frames
are sampled $\geq$60\,s away in the same video and keep their original
timestamp labels.}
\label{tab:app-corruption}
\end{table}

\begin{table}[tbh]
\centering
\small
\setlength{\tabcolsep}{5pt}
\begin{tabular}{lccc}
\toprule
\textbf{Oracle window dilation} & \eva{} & \videomind{} & \videor{} \\
\midrule
$2\times$ & 45.7 {\tiny[40.3, 51.2]} & 50.3 {\tiny[44.6, 55.9]} & 46.7 {\tiny[40.7, 52.4]} \\
$4\times$ & 46.7 {\tiny[41.6, 51.7]} & 46.7 {\tiny[41.1, 52.3]} & 44.3 {\tiny[39.0, 49.8]} \\
$8\times$ & 43.3 {\tiny[38.0, 48.6]} & 46.0 {\tiny[40.0, 52.0]} & 45.0 {\tiny[39.7, 50.3]} \\
$16\times$ & 38.7 {\tiny[33.6, 43.7]} & 45.3 {\tiny[39.6, 51.0]} & 39.3 {\tiny[34.0, 44.8]} \\
\bottomrule
\end{tabular}
\vspace{4pt}
\caption{Dilation ladder for all three agents ($n{=}300$): accuracy (\%) when
the oracle window is expanded about its centre. The plateau through $8\times$
replicates across architectures; only $16\times$ ($\approx$5\,min around a
10\,s clue) degrades.}
\label{tab:app-dilation}
\end{table}

\section{Audit Details}
\label{app:audit-details}

Table~\ref{tab:app-systems} lists the nine agent$\times$frame-budget
configurations used in the system-level audit
(Section~\ref{sec:audit-results}), with their causally measured cascade
sensitivity (oracle $-$ random, set S, $n{=}400$ items). The retained benchmark
scores were recomputed from the per-item outputs of the preliminary study; the
recomputation reproduces Table~\ref{tab:preliminary-run} exactly (e.g.,
\videomind{} at $\settingF{=}32$ gives VideoHallucer 76.25/64.58/46.83).

\begin{table}[tbh]
\centering
\small
\setlength{\tabcolsep}{5pt}
\begin{tabular}{lcc@{\hspace{2em}}lcc}
\toprule
\textbf{Configuration} & \textbf{Sens.\ (pp)} & \textbf{95\% CI} &
\textbf{Configuration} & \textbf{Sens.\ (pp)} & \textbf{95\% CI} \\
\midrule
\eva{} $\settingF{=}16$ & 16.8 & [11.0, 22.5] & \videomind{} $\settingF{=}64$ & 13.8 & [9.2, 18.4] \\
\eva{} $\settingF{=}32$ & 17.8 & [11.9, 23.5] & \videor{} $\settingF{=}16$ & 11.5 & [6.8, 16.3] \\
\eva{} $\settingF{=}64$ & 15.0 & [10.0, 20.1] & \videor{} $\settingF{=}32$ & 16.3 & [11.9, 20.7] \\
\videomind{} $\settingF{=}16$ & 15.8 & [10.5, 21.0] & \videor{} $\settingF{=}64$ & 15.5 & [10.5, 20.5] \\
\videomind{} $\settingF{=}32$ & 16.5 & [11.6, 21.4] & & & \\
\bottomrule
\end{tabular}
\vspace{4pt}
\caption{The nine configurations of the system-level audit and their causal
cascade sensitivity. Across these systems, the Spearman correlations between
each retained benchmark score and sensitivity are: LVBench-TG $-0.14$
[$-0.86, +0.79$], ARGUS ROUGE-L $-0.10$ [$-0.72, +0.75$], VideoHallucer basic
$+0.25$ [$-0.67, +0.74$], halluc $-0.02$ [$-0.63, +0.67$], both $-0.17$
[$-0.75, +0.60$] (95\% BCa); every interval spans zero.}
\label{tab:app-systems}
\end{table}

\section{Protocol Details}
\label{app:protocol-details}

\paragraph{EVA under the \condfull{} condition.}
Denied its frame-selection tool, \eva{} initially refused to answer 20\% of
items; a single forced-answer follow-up prompt reduced refusals to 4\%, and
every nudged run is flagged \texttt{forced\_answer} in the released logs.
Without this control, every \eva{} \condoracle{}$-$\condfull{} contrast would
be inflated by a format artefact rather than a grounding effect.

\paragraph{Multi-interval clues.}
11\% of items carry multiple ground-truth clue intervals; since all three
agents consume one contiguous window, these are served as their temporal hull.

\paragraph{Abstention.}
Unparseable output counts as wrong and is additionally reported as abstention;
abstention rates are $\leq$1.7\% in every grounding condition
(max: \eva{} \condoff{}).

\paragraph{Dr.V diagnosis coverage.}
Diagnosis failure rates are 3.3--16.7\% on \drvbench{} and 1.1--3.4\% on
\cgbench{}; rates and per-level (perceptive/temporal/cognitive) attributions
are included in the released reports. On \cgbench{} the graded verdict variant
is as flat as the binary one (0.746--0.768 across all cells).

\paragraph{Compute and reproduction.}
The campaign comprises 60{,}008 intervened runs and 5{,}784 diagnoses. Item
manifests and span generation are seeded, so all three agents receive
byte-identical spans and a re-run reproduces every number. Each run row records
its condition, requested and effective span, frame timestamps, frame-level
evidence rate, span IoU, parse path, and latency. Only the 227 required videos
(64\,GB) of the gated 411\,GB \cgbench{} release were fetched, by reading each
remote archive's central directory over HTTP range requests.

\section{Related Work}
\label{sec:related}

\subsection{Video Understanding Pipelines}

Video understanding has evolved from monolithic vision--language models that directly answer questions from uniformly sampled video frames to increasingly agentic systems that decompose reasoning into multiple stages~\cite{tang2023videounderstanding,maaz2024videochatgpt,zhang2023videollama,lin2024videollava,li2024llamavid,wang2024videoagentlong}.
Early video-language models typically process an entire video through a single forward pass and generate an answer without exposing intermediate reasoning, making their decision process difficult to interpret or diagnose~\cite{maaz2024videochatgpt,zhang2023videollama,lin2024videollava,li2024llamavid}.
To improve temporal grounding, long-video understanding, and reasoning efficiency, recent approaches increasingly decompose video understanding into multiple functional stages, including temporal localization, visual observation, memory retrieval, and answer synthesis~\cite{song2024moviechat,ren2024timechat,huang2024vtimellm,fan2024videoagentmemory,wang2024videoagentlong,liu2025videomind}.
Representative systems include retrieval-based pipelines, multi-agent frameworks, and role-specialized architectures, where different modules are responsible for locating relevant moments, extracting visual evidence, and integrating observations into a final response~\cite{wang2024videoagentlong,fan2024videoagentmemory,liu2025videomind,zhang2025deepvideodiscovery,liu2025longvideoagent,zhang2026eva}.
Such modular designs substantially improve scalability, interpretability, and long-video reasoning capability, and have rapidly become a common paradigm for modern video understanding systems~\cite{tang2023videounderstanding,wang2024videoagentlong,zhang2025deepvideodiscovery,liu2025longvideoagent,zhang2026eva}.
However, while video understanding has evolved into an inherently multi-stage process, its evaluation largely remains end-to-end~\cite{mangalam2023egoschema,li2024mvbench,fu2025videomme}. Existing studies primarily assess only the correctness of the final prediction, providing limited insight into how errors emerge and propagate across the underlying reasoning pipeline.

\subsection{Video Hallucination Evaluation}

Hallucination has recently emerged as one of the central challenges in video understanding, motivating a growing number of benchmark datasets and evaluation protocols~\cite{huang2026videohallucsurvey,wang2024videohallucer,zhang2024eventhallusion,kong2025mhbench,li2025vidhalluc,rawal2025argus}.
Existing benchmarks characterize hallucination from different perspectives, including visual fabrication, temporal inconsistency, semantic aggregation, descriptive omission, and event-level reasoning errors~\cite{wang2024videohallucer,yang2024vript,li2025vidhalluc,kong2025mhbench,rawal2025argus,zhang2024eventhallusion,lu2026elvhalluc}.
Some benchmarks evaluate whether generated answers are faithfully grounded in visual evidence, while others measure caption faithfulness or consistency between local observations and global event understanding~\cite{wang2024videohallucer,yang2024vript,rawal2025argus,zhang2024eventhallusion,lu2026elvhalluc}.
Together, these benchmarks have substantially advanced the systematic evaluation of hallucination in video-language models and enabled quantitative comparison across different architectures~\cite{huang2026videohallucsurvey,li2025vidhalluc,rawal2025argus,lu2026elvhalluc}.
Despite their different definitions and evaluation protocols, most existing benchmarks share a common assumption: hallucination is evaluated solely from the model's final output~\cite{wang2024videohallucer,li2025vidhalluc,rawal2025argus,lu2026elvhalluc}. Consequently, they determine whether an answer is hallucinated, but provide little evidence about where the hallucination originates.
For modern multi-stage video agents, however, hallucinated responses may arise from inaccurate temporal localization, incomplete visual observation, or incorrect reasoning over otherwise correct evidence. End-to-end evaluation therefore conflates failures from multiple reasoning stages, making targeted diagnosis and model improvement considerably more difficult.

\subsection{Benchmark Diagnosis}

Beyond proposing new benchmarks, an emerging line of research has begun to investigate the reliability and limitations of existing evaluation benchmarks themselves~\cite{chen2024rightway,yang2024mminstructeval,alzahrani2024whenbenchmarks,feng2025breakingdown,akhtar2026saturation}.
Previous studies have examined issues including benchmark saturation, dataset bias, inconsistent model rankings across benchmarks, shortcut learning, and the sensitivity of evaluation results to inference configurations~\cite{chen2024rightway,yang2024mminstructeval,alzahrani2024whenbenchmarks,ramos2025dataleakage,akhtar2026saturation,feng2025breakingdown,fu2026videommev2}.
These findings suggest that benchmark performance does not always faithfully reflect a model's underlying capability, motivating more reliable evaluation methodologies and benchmark diagnosis~\cite{chen2024rightway,alzahrani2024whenbenchmarks,akhtar2026saturation,feng2025breakingdown,fu2026videommev2}.
Our work is closely related to this direction but differs in both objective and methodology. Rather than introducing another hallucination benchmark or comparing benchmark difficulty, we investigate whether existing benchmarks remain informative when the evaluation target shifts from monolithic video-language models to modern multi-stage video agents.
Specifically, we organize representative benchmarks according to the functional stages of video understanding and study how effectively they diagnose hallucinations throughout the reasoning pipeline. Our analysis reveals that benchmarks developed independently for individual capabilities can produce inconsistent, incomplete, or even misleading signals when collectively evaluating modern agentic architectures.
Methodologically, prior benchmark-diagnosis studies remain observational: they compare scores across datasets, models, or inference settings, and thus inherit the confounds of the distributions they compare. We instead adopt the standard of \emph{controlled intervention}---overwriting a single pipeline stage with a known value while holding the downstream task fixed on one stage-annotated item set---which turns stage attribution from a correlational conjecture into a measured causal quantity, and provides a reference standard against which the predictive value of existing benchmark scores can itself be audited.

\section{EVA's Sweep on CG-Bench}

\begin{table*}[tbh]
\centering
\scriptsize
\setlength{\tabcolsep}{3.5pt}
\begin{tabular}{l|c|cccc|ccc}
\toprule
\multirow{2}{*}{\textbf{Method}} &
\multirow{2}{*}{\textbf{Size}} &
\multirow{2}{*}{\textbf{long-acc.}} &
\multirow{2}{*}{\textbf{mIoU}} &
\multirow{2}{*}{\textbf{rec.@IoU}} &
\multirow{2}{*}{\textbf{acc.@IoU}} &
\multicolumn{3}{c}{\textbf{Averaged Token Usage}} \\
\cmidrule(lr){7-9}
& & & & & & \textbf{Visual} & \textbf{Total} & \textbf{Vis.\ Frac.} \\
\midrule
EVA ($\mathit{mt}$=4, $\mathit{nf}$=16) & 7B & 36.93 & 4.41 & 4.97 & 2.63 & 8.1K & 9.2K & 85.8\% \\
EVA ($\mathit{mt}$=4, $\mathit{nf}$=32) & 7B & 39.00 & 4.47 & 4.90 & 2.57 & 13.2K & 14.7K & 88.3\% \\
EVA ($\mathit{mt}$=4, $\mathit{nf}$=64) & 7B & 41.00 & 4.83 & 5.35 & 2.69 & 16.5K & 18.5K & 88.3\% \\
\midrule
EVA ($\mathit{mt}$=6, $\mathit{nf}$=16) & 7B & 38.17 & 4.58 & 5.28 & \underline{2.93} & 8.2K & 9.5K & 84.5\% \\
EVA ($\mathit{mt}$=6, $\mathit{nf}$=32) & 7B & 40.20 & 4.51 & 4.99 & 2.59 & 13.3K & 14.9K & 87.8\% \\
EVA ($\mathit{mt}$=6, $\mathit{nf}$=64) & 7B & \textbf{41.70} & \textbf{4.87} & \underline{5.39} & 2.85 & 16.6K & 18.7K & 88.2\% \\
\midrule
EVA ($\mathit{mt}$=8, $\mathit{nf}$=16) & 7B & 37.53 & 4.51 & 5.28 & \textbf{3.17} & 8.1K & 9.4K & 84.4\% \\
EVA ($\mathit{mt}$=8, $\mathit{nf}$=32) & 7B & 40.00 & 4.64 & 5.13 & 2.71 & 13.3K & 14.9K & 87.9\% \\
EVA ($\mathit{mt}$=8, $\mathit{nf}$=64) & 7B & 40.70 & 4.77 & 5.38 & 2.69 & 16.6K & 18.6K & 88.2\% \\
\midrule
EVA ($\mathit{mt}$=10, $\mathit{nf}$=16) & 7B & 37.77 & 4.50 & 5.14 & 2.83 & 8.2K & 9.5K & 84.6\% \\
EVA ($\mathit{mt}$=10, $\mathit{nf}$=32) & 7B & 39.87 & 4.52 & 4.87 & 2.58 & 13.2K & 14.8K & 87.8\% \\
EVA ($\mathit{mt}$=10, $\mathit{nf}$=64) & 7B & \underline{41.03} & 4.79 & 5.32 & 2.61 & 16.6K & 18.7K & 88.1\% \\
\midrule
EVA ($\mathit{mt}$=16, $\mathit{nf}$=16) & 7B & 37.83 & 4.49 & 5.21 & 2.73 & 8.2K & 9.5K & 84.5\% \\
EVA ($\mathit{mt}$=16, $\mathit{nf}$=32) & 7B & 39.37 & 4.64 & 5.21 & 2.57 & 13.3K & 14.9K & 87.8\% \\
EVA ($\mathit{mt}$=16, $\mathit{nf}$=64) & 7B & 40.80 & 4.77 & 5.27 & 2.64 & 16.8K & 18.9K & 88.2\% \\
\bottomrule
\end{tabular}

\caption{Performance of EVA variants on CG-Bench \citep{chen2025cg}.
$\mathit{mt}$ denotes the maximum number of tool-call turns and
$\mathit{nf}$ the number of frames per tool call.}
\label{tab:cgbench_ablation}
\end{table*}

\section{Eva's Sweep on Next-GQA}

\begin{table*}[ht]
\centering
\scriptsize
\setlength{\tabcolsep}{3pt}
\begin{tabular}{l|c|ccc|ccc|cc|ccc}
\toprule
\multirow{2}{*}{\textbf{Method}} &
\multirow{2}{*}{\textbf{Size}} &
\multicolumn{3}{c|}{\textbf{IoU}} &
\multicolumn{3}{c|}{\textbf{IoP}} &
\multirow{2}{*}{\textbf{Long Acc}} &
\multirow{2}{*}{\textbf{Acc@GQA}} &
\multicolumn{3}{c}{\textbf{Averaged Token Usage}} \\
\cmidrule(lr){3-5}
\cmidrule(lr){6-8}
\cmidrule(lr){11-13}
&& R@0.3 & R@0.5 & mIoU
& R@0.3 & R@0.5 & mIoP
&&
& \textbf{Visual} & \textbf{Total} & \textbf{Vis.\ Frac.} \\
\midrule
EVA ($\mathit{mt}$=4, $\mathit{nf}$=16)  & 7B & \underline{35.47} & 17.46 & \underline{27.34} & \textbf{37.44} & 20.17 & \textbf{29.83} & \underline{69.43} & 14.84 & 3.2K & 4.0K & 79.9\% \\
EVA ($\mathit{mt}$=4, $\mathit{nf}$=32)  & 7B & 34.51 & 16.60 & 26.93 & 36.10 & 19.20 & 29.21 & 68.99 & 14.26 & 5.9K & 7.0K & 84.8\% \\
EVA ($\mathit{mt}$=4, $\mathit{nf}$=64)  & 7B & 33.46 & 16.35 & 26.62 & 34.84 & 18.59 & 28.69 & 68.95 & 13.80 & 10.1K & 11.6K & 86.7\% \\
\midrule
EVA ($\mathit{mt}$=6, $\mathit{nf}$=16)  & 7B & 35.37 & 17.25 & 27.29 & 37.04 & 20.06 & \underline{29.82} & 69.25 & 14.76 & 3.2K & 4.0K & 80.0\% \\
EVA ($\mathit{mt}$=6, $\mathit{nf}$=32)  & 7B & 34.55 & 16.39 & 26.87 & 36.16 & 19.14 & 29.22 & 69.27 & 13.92 & 5.9K & 6.9K & 84.7\% \\
EVA ($\mathit{mt}$=6, $\mathit{nf}$=64)  & 7B & 33.38 & 16.25 & 26.52 & 34.91 & 18.62 & 28.62 & 68.87 & 13.58 & 10.1K & 11.7K & 86.6\% \\
\midrule
EVA ($\mathit{mt}$=8, $\mathit{nf}$=16)  & 7B & 35.37 & 17.40 & 27.27 & \underline{37.15} & 20.08 & 29.68 & 68.97 & 14.72 & 3.2K & 4.0K & 79.8\% \\
EVA ($\mathit{mt}$=8, $\mathit{nf}$=32)  & 7B & 34.63 & 16.67 & 27.02 & 36.33 & 19.31 & 29.30 & 68.85 & 14.19 & 5.8K & 6.9K & 84.6\% \\
EVA ($\mathit{mt}$=8, $\mathit{nf}$=64)  & 7B & 33.37 & 16.27 & 26.61 & 34.88 & 18.55 & 28.73 & 69.31 & 13.42 & 10.1K & 11.7K & 86.7\% \\
\midrule
EVA ($\mathit{mt}$=10, $\mathit{nf}$=16) & 7B & \textbf{35.66} & \textbf{17.55} & \textbf{27.39} & \textbf{37.44} & \textbf{20.29} & \textbf{29.83} & 69.16 & \textbf{15.01} & 3.2K & 4.0K & 79.8\% \\
EVA ($\mathit{mt}$=10, $\mathit{nf}$=32) & 7B & 34.23 & 16.73 & 26.90 & 35.95 & 19.24 & 29.17 & 69.37 & 14.09 & 5.9K & 6.9K & 84.6\% \\
EVA ($\mathit{mt}$=10, $\mathit{nf}$=64) & 7B & 33.17 & 16.21 & 26.52 & 34.67 & 18.59 & 28.65 & 68.32 & 13.61 & 10.2K & 11.7K & 86.7\% \\
\midrule
EVA ($\mathit{mt}$=16, $\mathit{nf}$=16) & 7B & 35.24 & \underline{17.51} & 27.31 & 37.04 & \underline{20.23} & 29.70 & \underline{69.43} & \underline{14.99} & 3.2K & 4.0K & 79.9\% \\
EVA ($\mathit{mt}$=16, $\mathit{nf}$=32) & 7B & 34.53 & 16.86 & 27.03 & 36.12 & 19.50 & 29.36 & \textbf{69.90} & 14.28 & 5.9K & 6.9K & 84.6\% \\
EVA ($\mathit{mt}$=16, $\mathit{nf}$=64) & 7B & 33.38 & 16.18 & 26.56 & 34.88 & 18.55 & 28.63 & 68.89 & 13.54 & 10.1K & 11.6K & 86.7\% \\
\bottomrule
\end{tabular}

\caption{Ablation study on EVA sweep configurations on NExT-GQA \citep{nextgqa}.
$\mathit{mt}$ denotes the maximum number of tool-call turns and
$\mathit{nf}$ the number of frames per tool call.}
\label{tab:nextgqa_ablation}
\end{table*}

\section{LVBench and ARGUS Performance}
\label{app:argus}

\begin{table*}[ht]
	\centering
	\small
	\setlength{\tabcolsep}{5pt}
	\begin{tabular}{lcccccc}
		\toprule
		\textbf{Model} &
		\textbf{Entity} &
		\textbf{Event} &
		\textbf{Key Info.} &
		\textbf{Reasoning} &
		\textbf{Summarization} &
		\textbf{Overall} \\
		&
		\textbf{Recognition} &
		\textbf{Understanding} &
		\textbf{Retrieval} &
		&
		&
		\textbf{Avg.} \\
		\midrule
		VideoMind-7B       & 62.90 & 57.61 & 62.50 & 29.73 & 54.14 &  54.79 \\
		VideoMind-2B       & 61.29 & 58.70 & 58.33 & 37.84 & 50.00 & 55.25 \\
		Video-R1                & 58.06 & 65.22 & 66.67 & 45.95 & 42.86 & 59.82 \\
		MARC-3B               & 38.71 & 30.43 & 58.33 & 29.73 & 14.29 & 35.16 \\
		EVA                         & 66.13 & 56.52 & 70.83 & 54.05 & 57.14 & 60.27 \\
		\midrule
		\# of samples         & 62 & 92 & 24 & 37 & 14 & total: 219 \\
		\bottomrule
	\end{tabular}
	\caption{Performance comparison on different LVBench categories. Metric: Accuracy (\%).}
	\label{tab:lvbench_categories}
\end{table*}

\begin{table}[ht]
	\centering
	\small
	\setlength{\tabcolsep}{4pt}
	\begin{tabular}{llcccc}
		\toprule
		\multirow{2}{*}{\textbf{Model}} &
		\multicolumn{2}{c}{\textbf{Param.}} &
		\multirow{2}{*}{$\mathbf{Cost_H}\downarrow$} &
		\multirow{2}{*}{$\mathbf{Cost_O}\downarrow$} \\
		\cmidrule(lr){2-3}
		& $\mathbf{n_{frame}}$ & $\mathbf{n_{max-tool}}$ & & \\
		\midrule
		
		\multirow{3}{*}{Video-R1}
		& 16 & -- & 0.6825 & 0.8384 \\
		& 32 & -- & 0.6615 & 0.8358 \\
		& 64 & -- & 0.6651 & 0.8328 \\
		\midrule
		
		\multirow{3}{*}{VideoMind-7B}
		& 16 & -- & 0.5518 & 0.8299 \\
		& 32 & -- & 0.5563 & 0.8306 \\
		& 64 & -- & 0.5540 & 0.8424 \\
		\midrule
		
		\multirow{15}{*}{EVA}
		& 16 & 4  & 0.4956 & 0.8254 \\
		& 32 & 4  & 0.4880 & 0.8157 \\
		& 64 & 4  & 0.4795 & 0.8141 \\
		
		& 16 & 6  & 0.5165 & 0.8262 \\
		& 32 & 6  & 0.5310 & 0.8281 \\
		& 64 & 6  & 0.5305 & 0.8159 \\
		
		& 16 & 8  & 0.5292 & 0.8190 \\
		& 32 & 8  & 0.5305 & 0.8156 \\
		& 64 & 8  & 0.5232 & 0.8107 \\
		
		& 16 & 10 & 0.5235 & 0.8254 \\
		& 32 & 10 & 0.5251 & 0.8191 \\
		& 64 & 10 & 0.5390 & 0.8136 \\
		
		& 16 & 16 & 0.5253 & 0.8184 \\
		& 32 & 16 & 0.5219 & 0.8199 \\
		& 64 & 16 & 0.5150 & 0.8223 \\
		\bottomrule
	\end{tabular}
    \vspace{4pt}
	\caption{Cost comparison under different frame numbers ($n_{frame}$) and maximum tool budgets ($max_{tool}$). Lower is better.}
	\label{tab:cost_comparison}
\end{table}

\section{Per-Sample Cascade Analysis: VideoMind Grounding Quality on CGBench}
\label{app:cascade_per_sample}

Table~\ref{tab:cascade_per_sample} reports how \videomind{}-7B's per-sample
grounding quality (measured as IoU between the Grounder's predicted span
and the ground-truth temporal span) correlates with final answer correctness
across $N=3{,}000$ CGBench questions.
The threshold IoU~$\geq 0.1$ is used to distinguish samples where the
Grounder at least partially localized the relevant segment from those where
it missed entirely; mean grounding IoU is 0.065, confirming that accurate
temporal localization is rare and constitutes the primary bottleneck.

\begin{table}[tbh]
\centering
\small
\setlength{\tabcolsep}{5pt}
\begin{tabular}{lrrrrrr}
\toprule
\textbf{Category} & \textbf{N} & \textbf{Mean IoU} &
\textbf{Acc\,(IoU$<$0.1)$\uparrow$} &
\textbf{Acc\,(IoU$\geq$0.1)$\uparrow$} &
\textbf{$\Delta$$\uparrow$} \\
\midrule
2D Spatial Perception   & 180  & 0.052 & 33.1 & 46.2 & +13.1 \\
Entity Cognition        & 300  & 0.046 & 27.5 & 23.8 & \phantom{+}$-$3.7 \\
Entity Perception       & 600  & 0.071 & 31.0 & 38.1 & \phantom{+}+7.0 \\
Event Cognition         & 300  & 0.087 & 40.3 & 51.4 & +11.1 \\
Event Perception        & 600  & 0.069 & 30.6 & 43.0 & +12.4 \\
Hallucination           & 240  & 0.054 & 38.8 & 59.0 & +20.2 \\
Scene Cognition         &  15  & 0.176 & 20.0 & 60.0 & +40.0 \\
Scene Perception        & 180  & 0.047 & 47.4 & 73.1 & +25.7 \\
Text Cognition          &  45  & 0.081 & 30.6 & 55.6 & +25.0 \\
Text Perception         & 300  & 0.065 & 44.4 & 54.0 & \phantom{+}+9.6 \\
Time Cognition          &  60  & 0.171 & 25.6 & 14.3 & \phantom{+}$-$11.4 \\
Time Perception         & 180  & 0.031 & 33.8 & 50.0 & +16.2 \\
\midrule
\textbf{Overall}        & \textbf{3000} & \textbf{0.065} & \textbf{34.6} & \textbf{44.9} & \textbf{+10.3} \\
\bottomrule
\end{tabular}
\vspace{4pt}
\caption{Per-sample cascade analysis for \videomind{}-7B on CGBench
(default configuration, $\mathit{nf}=32$).
For each CGBench question the Grounder's predicted temporal span is
compared against the ground-truth span to compute IoU.
``Acc\,(IoU$<$0.1)'' and ``Acc\,(IoU$\geq$0.1)'' are final answer
accuracies when grounding fails and succeeds, respectively.
$\Delta$ is the cascade lift in percentage points.
Scene, Hallucination, and Text categories show the strongest cascade
($>$20pp); Time Cognition reverses, suggesting temporal-reasoning
questions are impaired by narrow window grounding.
$r = 0.072$, $p < 0.001$ overall.}
\label{tab:cascade_per_sample}
\end{table}

\section{Performance on ELV-Halluc Dataset}

	\begin{table*}[ht]
		\centering
		\footnotesize
		\setlength{\tabcolsep}{2pt}
		
		\begin{tabular}{lcccccccccccccccc}
			\toprule
			
			\multirow{2}{*}{\textbf{Models}} &
			\multirow{2}{*}{\makecell{\textbf{LLM}\\\textbf{Size}}} &
			\multicolumn{3}{c}{\textbf{Visual Details}} &
			\multicolumn{3}{c}{\textbf{Object}} &
			\multicolumn{3}{c}{\textbf{Action}} &
			\multicolumn{3}{c}{\textbf{Declarative Content}} &
			\multirow{2}{*}{\makecell{\textbf{Avg}\\\textbf{Acc}$\uparrow$}} &
			\multirow{2}{*}{\makecell{\textbf{Avg}\\\textbf{Diff.}$\downarrow$}} &
			\multirow{2}{*}{\makecell{\textbf{SAH}\\\textbf{Ratio}$\downarrow$}}
			\\
			
			\cmidrule(lr){3-5}
			\cmidrule(lr){6-8}
			\cmidrule(lr){9-11}
			\cmidrule(lr){12-14}
			
			&
			&
			\textbf{In.} & \textbf{Out.} & \textbf{Diff.}
			&
			\textbf{In.} & \textbf{Out.} & \textbf{Diff.}
			&
			\textbf{In.} & \textbf{Out.} & \textbf{Diff.}
			&
			\textbf{In.} & \textbf{Out.} & \textbf{Diff.}
			&
			&
			&
			\\
			
			\midrule
			
			\multicolumn{17}{c}{\textbf{Existing Results from ELV-Halluc}} \\
			\midrule
			
			InternVL3-1B
			& 0.5B
			& 8 & 11 & 3
			& 8.7 & 11 & 2.3
			& 8.7 & 12.5 & 3.8
			& 11.3 & 8.3 & -3
			& 9.9 & 1.5 & 1.6
			\\
			
			InternVL3-2B
			& 1.5B
			& 7 & 15.5 & 8.5
			& 8.7 & 17.2 & 8.5
			& 7.2 & 10.5 & 3.3
			& 10 & 13 & 3
			& 11.1 & 5.8 & 6.3
			\\
			
			SmolVLM-2.2B
			& 1.7B
			& 0 & 0 & 0
			& 3 & 5 & 2
			& 0 & 0 & 0
			& 0 & 0 & 0
			& 1 & 0.5 & 0.5
			\\
			
			Qwen2.5VL-3B
			& 3B
			& 2.2 & 10.5 & 8.3
			& 7.7 & 13.8 & 6.1
			& 5 & 8 & 3
			& 6 & 6 & 0
			& 7.4 & 4.3 & 4.5
			\\
			
			LLaVA-Video-7B
			& 7B
			& 3.7 & 3.7 & 0
			& 4.5 & 2.5 & -2
			& 3.7 & 3.2 & -0.5
			& 4 & 4 & 0
			& 3.6 & -0.6 & -0.6
			\\
			
			Video-chatgpt-7B
			& 7B
			& 2 & 2.5 & 0.5
			& 2.5 & 1.7 & -0.7
			& 1.2 & 1.2 & 0
			& 2.2 & 3.2 & 1
			& 2.0 & 0.2 & 0.1
			\\
			
			LLaVA-OV-7B
			& 7B
			& 8 & 13.2 & 5.2
			& 9.5 & 13.7 & 4.2
			& 8.7 & 10.7 & 2
			& 7.7 & 7.5 & -0.2
			& 9.9 & 2.8 & 3.0
			\\
			
			Qwen2.5VL-7B
			& 7B
			& 10.2 & 26 & 15.8
			& 17.5 & 30.7 & 13.2
			& 13 & 20.7 & 7.7
			& 16.8 & 10.5 & -6.3
			& 18.1 & 7.6 & 8.8
			\\
			
			InternVL3-8B
			& 7B
			& 12.5 & 19.5 & 7.0
			& 14.5 & 19.5 & 5.0
			& 13.5 & 20.5 & 7.0
			& 12.8 & 17.7 & 4.9
			& 16.3 & 5.9 & 6.8
			\\
			
			InternVL3-14B
			& 14B
			& 17.5 & 24.5 & 7.0
			& 22.8 & 24.5 & 1.7
			& 16.3 & 17.7 & 1.4
			& 15.2 & 15.5 & 0.3
			& 19.2 & 2.6 & 3.1
			\\
			
			Qwen2.5VL-32B
			& 32B
			& 16.5 & 24.5 & 8.0
			& 21.7 & 24.5 & 2.8
			& 17.2 & 15.0 & -2.2
			& 15.2 & 7.2 & -8.0
			& 17.7 & 0.1 & 0.2
			\\
			
			InternVL3-38B
			& 32B
			& 25.3 & 29 & 3.7
			& 24.2 & 28 & 3.8
			& 24 & 30 & 6
			& 24.5 & 24.2 & -0.3
			& 26.1 & 3.3 & 4.3
			\\
			
			Qwen2.5VL-72B
			& 72B
			& 24 & 35.5 & 11.5
			& 35.7 & 41.5 & 5.8
			& 27.8 & 32.3 & 4.5
			& 32.3 & 27 & -5.3
			& 32.0 & 4.1 & 5.8
			\\
			
			InternVL3-78B
			& 72B
			& 25 & 31.2 & 6.2
			& 32 & 36.5 & 4.5
			& 28.5 & 31.2 & 2.7
			& 24.2 & 26.5 & 2.3
			& 29.3 & 3.9 & 5.4
			\\

			GPT-4o
			& /
			& 7.7 & 8.3 & 0.6
			& 8 & 8.7 & 0.7
			& 8.7 & 10.2 & 1.5
			& 8.5 & 9.5 & 1
			& 8.7 & 0.9 & 1.0
			\\
			
			Gemini2.5-Flash
			& /
			& 47 & 58 & 11
			& 56.5 & 58.8 & 2.3
			& 50.5 & 53.2 & 2.7
			& 48.7 & 52 & 3.3
			& 53.1 & 4.8 & 9.8
			\\
			
			\midrule

		\multicolumn{17}{c}{\textbf{VideoMind}} \\
		\midrule
		
		VideoMind-2B (f=16)
		& 2B
		& 9.0 & 11.0 & 2.1
		& 8.2 & 9.7 & 1.5
		& 10.5 & 11.8 & 1.3
		& 9.5 & 7.9 & -1.5
		& 9.7 & 0.8 & 0.9
		\\
		
		VideoMind-2B (f=32)
		& 2B
		& 9.7 & 11.5 & 1.8
		& 7.7 & 9.2 & 1.5
		& 9.5 & 11.0 & 1.5
		& 8.7 & 8.7 & 0.0
		& 9.5 & 1.2 & 1.3
		\\
		
		VideoMind-7B (f=16)
		& 7B
		& 12.6 & 20.0 & 7.4
		& 18.2 & 24.6 & 6.4
		& 13.1 & 15.1 & 2.1
		& 13.8 & 12.6 & -1.3
		& 16.2 & 3.7 & 4.3
		\\
		
		VideoMind-7B (f=32)
		& 7B
		& 6.9 & 14.9 & 8.0
		& 12.1 & 19.5 & 7.4
		& 10.0 & 10.5 & 0.5
		& 8.2 & 9.0 & 0.8
		& 11.4 & 4.2 & 4.6
		\\
		
		VideoMind-2B (f=64)
		& 2B
		& 13.8 & 13.6 & -0.3
		& 12.3 & 12.3 & 0.0
		& 11.3 & 13.6 & 2.3
		& 9.7 & 9.0 & -0.8
		& 12.0 & 0.3 & 0.4
		\\
		
		VideoMind-7B (f=64)
		& 7B
		& 6.2 & 10.0 & 3.8
		& 9.0 & 16.2 & 7.2
		& 6.7 & 10.3 & 3.6
		& 5.6 & 7.2 & 1.5
		& 8.9 & 4.0 & 4.3
		\\
		
		VideoMind-2B (f=128)
		& 2B
		& 21.3 & 20.5 & -0.8
		& 17.7 & 19.7 & 2.1
		& 21.0 & 20.8 & -0.3
		& 21.0 & 19.7 & -1.3
		& 20.2 & -0.1 & -0.1
		\\
		
		VideoMind-7B (f=128)
		& 7B
		& 3.6 & 8.2 & 4.6
		& 6.9 & 10.3 & 3.3
		& 4.6 & 8.5 & 3.8
		& 5.6 & 5.4 & -0.3
		& 6.6 & 2.9 & 3.0
		\\
		
		VideoMind-2B (f=256)
		& 2B
		& 25.6 & 29.5 & 3.8
		& 25.1 & 24.1 & -1.0
		& 27.7 & 28.2 & 0.5
		& 26.2 & 24.6 & -1.5
		& 26.4 & 0.4 & 0.6
		\\
		
		VideoMind-7B (f=256)
		& 7B
		& 2.8 & 5.1 & 2.3
		& 4.4 & 8.5 & 4.1
		& 3.3 & 5.4 & 2.1
		& 4.1 & 4.4 & 0.3
		& 4.7 & 2.2 & 2.3
		\\
		
		\midrule
		\multicolumn{17}{c}{\textbf{Video-R1 (frame-count sweep)}} \\
		\midrule
		Video-R1-7B-16F
		& 7B
		& 12.1 & 15.9 & 3.8
		& 14.9 & 15.4 & 0.5
		& 15.1 & 13.1 & -2.1
		& 19.2 & 15.9 & -3.3
		& 15.2 & -0.3 & -0.3
		\\

		Video-R1-7B-32F
		& 7B
		& 9.2 & 15.9 & 6.7
		& 10.3 & 17.7 & 7.4
		& 13.3 & 16.2 & 2.8
		& 13.3 & 13.3 & 0.0
		& 13.7 & 4.2 & 4.8
		\\

		Video-R1-7B-64F
		& 7B
		& 6.9 & 13.8 & 6.9
		& 10.0 & 15.9 & 5.9
		& 10.0 & 14.9 & 4.9
		& 12.3 & 8.2 & -4.1
		& 11.5 & 3.4 & 3.8
		\\

		\midrule
		\multicolumn{17}{c}{\textbf{EVA (parameter sweep)}} \\
		\midrule
		EVA (mt4, 16F)
		& 7B
		& 0.5 & 0.8 & 0.3
		& 0.0 & 1.8 & 1.8
		& 0.8 & 0.3 & -0.5
		& 0.5 & 0.0 & -0.5
		& 0.6 & 0.3 & 0.3
		\\

		EVA (mt4, 32F)
		& 7B
		& 0.3 & 0.3 & 0.0
		& 1.5 & 1.0 & -0.5
		& 0.5 & 0.0 & -0.5
		& 1.0 & 0.0 & -1.0
		& 0.6 & -0.5 & -0.5
		\\

		EVA (mt4, 64F)
		& 7B
		& 0.8 & 0.5 & -0.3
		& 0.5 & 0.5 & 0.0
		& 0.5 & 1.0 & 0.5
		& 0.5 & 0.0 & -0.5
		& 0.5 & -0.1 & -0.1
		\\

		EVA (mt4, 128F)
		& 7B
		& 0.3 & 0.3 & 0.0
		& 0.5 & 0.5 & 0.0
		& 0.5 & 0.3 & -0.3
		& 0.5 & 0.0 & -0.5
		& 0.4 & -0.2 & -0.2
		\\

		EVA (mt6, 16F)
		& 7B
		& 0.0 & 1.0 & 1.0
		& 0.5 & 1.5 & 1.0
		& 0.3 & 0.3 & 0.0
		& 0.8 & 0.0 & -0.8
		& 0.5 & 0.3 & 0.3
		\\

		EVA (mt6, 32F)
		& 7B
		& 0.5 & 0.8 & 0.3
		& 1.3 & 0.3 & -1.0
		& 0.5 & 0.3 & -0.3
		& 0.5 & 0.3 & -0.3
		& 0.5 & -0.3 & -0.3
		\\

		EVA (mt6, 64F)
		& 7B
		& 0.3 & 0.8 & 0.5
		& 0.3 & 0.8 & 0.5
		& 0.3 & 0.5 & 0.3
		& 0.8 & 0.0 & -0.8
		& 0.4 & 0.1 & 0.1
		\\

		EVA (mt6, 128F)
		& 7B
		& 0.3 & 0.8 & 0.5
		& 0.8 & 0.8 & 0.0
		& 0.5 & 0.8 & 0.3
		& 0.0 & 0.3 & 0.3
		& 0.5 & 0.3 & 0.3
		\\

		EVA (mt8, 16F)
		& 7B
		& 0.0 & 1.0 & 1.0
		& 1.3 & 1.3 & 0.0
		& 0.5 & 0.8 & 0.3
		& 0.5 & 0.0 & -0.5
		& 0.7 & 0.2 & 0.2
		\\

		EVA (mt8, 32F)
		& 7B
		& 0.3 & 0.3 & 0.0
		& 0.8 & 0.5 & -0.3
		& 0.5 & 0.3 & -0.3
		& 0.8 & 0.0 & -0.8
		& 0.4 & -0.3 & -0.3
		\\

		EVA (mt8, 64F)
		& 7B
		& 0.3 & 0.5 & 0.3
		& 1.0 & 0.8 & -0.3
		& 0.3 & 0.8 & 0.5
		& 0.3 & 0.0 & -0.3
		& 0.5 & 0.1 & 0.1
		\\

		EVA (mt8, 128F)
		& 7B
		& 0.3 & 0.0 & -0.3
		& 0.5 & 0.3 & -0.3
		& 0.3 & 0.5 & 0.3
		& 0.0 & 0.0 & 0.0
		& 0.2 & -0.1 & -0.1
		\\

		EVA (mt10, 16F)
		& 7B
		& 0.3 & 1.5 & 1.3
		& 1.0 & 1.3 & 0.3
		& 0.5 & 0.8 & 0.3
		& 0.8 & 0.0 & -0.8
		& 0.8 & 0.3 & 0.3
		\\

		EVA (mt10, 32F)
		& 7B
		& 0.3 & 0.8 & 0.5
		& 1.3 & 1.0 & -0.3
		& 0.3 & 0.5 & 0.3
		& 0.8 & 0.3 & -0.5
		& 0.6 & 0.0 & 0.0
		\\

		EVA (mt10, 64F)
		& 7B
		& 0.3 & 1.0 & 0.8
		& 0.5 & 0.8 & 0.3
		& 0.0 & 1.0 & 1.0
		& 0.5 & 0.0 & -0.5
		& 0.5 & 0.4 & 0.4
		\\

		EVA (mt10, 128F)
		& 7B
		& 0.0 & 0.0 & 0.0
		& 0.3 & 0.3 & 0.0
		& 0.0 & 0.3 & 0.3
		& 0.5 & 0.0 & -0.5
		& 0.2 & -0.1 & -0.1
		\\

		EVA (mt16, 16F)
		& 7B
		& 0.5 & 1.3 & 0.8
		& 1.0 & 1.0 & 0.0
		& 0.8 & 0.5 & -0.3
		& 0.3 & 0.3 & 0.0
		& 0.7 & 0.1 & 0.1
		\\

		EVA (mt16, 32F)
		& 7B
		& 0.3 & 0.5 & 0.3
		& 0.5 & 1.0 & 0.5
		& 0.5 & 0.5 & 0.0
		& 0.5 & 0.3 & -0.3
		& 0.5 & 0.1 & 0.1
		\\

		EVA (mt16, 64F)
		& 7B
		& 0.8 & 1.0 & 0.3
		& 1.0 & 1.0 & 0.0
		& 0.0 & 0.8 & 0.8
		& 0.5 & 0.0 & -0.5
		& 0.6 & 0.1 & 0.1
		\\

			\bottomrule
		\end{tabular}
		
		\caption{
			Performance on ELV-Halluc.
			``In.'' and ``Out.'' denote accuracies on in-video and out-video hallucination samples, respectively.
			``Diff.'' denotes the difference between the two accuracies.
			Lower SAH Ratio indicates better robustness against semantic aggregation hallucinations.
		}
		\label{tab:elv_halluc}
	\end{table*}

\section{Performance on VideoHallucer}
\label{app:videohallucer}

These tables report per-category performance on VideoHallucer \citep{videohallucer2024} for Video-R1, VideoMind-7B, and EVA under all evaluated configurations.
$\mathit{nf}$ denotes the number of input frames; for EVA, $\mathit{mt}$ additionally denotes the maximum number of tool-call turns.
Three metrics are reported: \textit{Basic Acc} (accuracy on basic questions), \textit{Halluc Acc} (accuracy on hallucination questions), and \textit{Both Acc} (both questions simultaneously correct).

\begin{table*}[ht]
\centering
\scriptsize
\setlength{\tabcolsep}{4pt}
\begin{tabularx}{\linewidth}{l|c|c|ccccccc}
\toprule
\textbf{Method} & \textbf{Size} & \textbf{Overall} & \textbf{Obj-Rel} & \textbf{Temporal} & \textbf{Sem-Det} & \textbf{Ext-Fact} & \textbf{Ext-NF} & \textbf{Fact-Det} & \textbf{Interact} \\
\midrule
Video-R1 ($\mathit{nf}$=16) & 7B & 81.08 & 72.00 & 55.11 & 93.00 & 91.50 & 89.50 & 98.00 & 69.35 \\
Video-R1 ($\mathit{nf}$=32) & 7B & 71.67 & 28.00 & 51.70 & 84.00 & 91.50 & 92.00 & 94.00 & 67.74 \\
Video-R1 ($\mathit{nf}$=64) & 7B & 64.08 &  4.50 & 60.80 & 52.00 & 91.00 & 93.00 & 94.00 & 70.16 \\
\midrule
VideoMind ($\mathit{nf}$=16) & 7B & 74.75 & 81.00 & 55.68 & 89.00 & 89.50 & 89.50 & 34.00 & 54.03 \\
VideoMind ($\mathit{nf}$=32) & 7B & 76.25 & 82.50 & 56.25 & 92.00 & 90.00 & 90.00 & 38.00 & 55.65 \\
VideoMind ($\mathit{nf}$=64) & 7B & 79.00 & 82.00 & 61.93 & 88.50 & 92.50 & 92.50 & 60.00 & 54.84 \\
\midrule
EVA ($\mathit{mt}$=4,  $\mathit{nf}$=16)  & 7B & 88.50 & 84.00 & 91.48 & 93.00 & 92.00 & 93.00 & 98.00 & 63.71 \\
EVA ($\mathit{mt}$=4,  $\mathit{nf}$=32)  & 7B & 88.25 & 84.50 & 94.89 & 92.00 & 94.00 & 89.00 & 98.00 & 60.48 \\
EVA ($\mathit{mt}$=4,  $\mathit{nf}$=64)  & 7B & 88.00 & 82.00 & 95.45 & 91.50 & 92.00 & 92.50 & 99.00 & 58.87 \\
EVA ($\mathit{mt}$=4,  $\mathit{nf}$=128) & 7B & 87.58 & 81.50 & 94.32 & 93.50 & 91.50 & 90.50 & 98.00 & 58.87 \\
\midrule
EVA ($\mathit{mt}$=6,  $\mathit{nf}$=16)  & 7B & 88.42 & 85.50 & 93.18 & 92.50 & 92.50 & 92.50 & 97.00 & 59.68 \\
EVA ($\mathit{mt}$=6,  $\mathit{nf}$=32)  & 7B & 87.17 & 82.50 & 92.05 & 91.50 & 91.50 & 89.50 & 98.00 & 61.29 \\
EVA ($\mathit{mt}$=6,  $\mathit{nf}$=64)  & 7B & 88.00 & 83.00 & 93.18 & 92.50 & 93.50 & 90.50 & 98.00 & 60.48 \\
EVA ($\mathit{mt}$=6,  $\mathit{nf}$=128) & 7B & 88.25 & 80.50 & 94.32 & 93.00 & 92.00 & 92.50 & 99.00 & 62.90 \\
\midrule
EVA ($\mathit{mt}$=8,  $\mathit{nf}$=16)  & 7B & 88.08 & 82.50 & 93.18 & 93.00 & 92.50 & 91.00 & 97.00 & 62.90 \\
EVA ($\mathit{mt}$=8,  $\mathit{nf}$=32)  & 7B & 88.42 & 85.00 & 96.02 & 92.00 & 90.50 & 91.00 & 100.00 & 60.48 \\
EVA ($\mathit{mt}$=8,  $\mathit{nf}$=64)  & 7B & 88.83 & 83.00 & 95.45 & 92.50 & 92.50 & 92.50 & 99.00 & 62.90 \\
EVA ($\mathit{mt}$=8,  $\mathit{nf}$=128) & 7B & 87.92 & 83.50 & 92.61 & 94.00 & 90.00 & 92.50 & 98.00 & 59.68 \\
\midrule
EVA ($\mathit{mt}$=10, $\mathit{nf}$=16)  & 7B & 88.25 & 81.00 & 93.18 & 94.00 & 91.50 & 92.50 & 99.00 & 62.90 \\
EVA ($\mathit{mt}$=10, $\mathit{nf}$=32)  & 7B & 87.67 & 82.00 & 94.89 & 91.50 & 92.50 & 91.00 & 99.00 & 58.06 \\
EVA ($\mathit{mt}$=10, $\mathit{nf}$=64)  & 7B & 88.00 & 83.00 & 93.75 & 93.50 & 92.50 & 90.50 & 98.00 & 59.68 \\
EVA ($\mathit{mt}$=10, $\mathit{nf}$=128) & 7B & 88.33 & 82.00 & 94.89 & 93.00 & 93.00 & 91.00 & 99.00 & 61.29 \\
\midrule
EVA ($\mathit{mt}$=16, $\mathit{nf}$=16)  & 7B & 88.42 & 79.50 & 94.89 & 94.00 & 93.50 & 92.00 & 98.00 & 62.90 \\
EVA ($\mathit{mt}$=16, $\mathit{nf}$=32)  & 7B & 87.17 & 81.00 & 92.05 & 92.00 & 91.50 & 91.00 & 100.00 & 58.87 \\
EVA ($\mathit{mt}$=16, $\mathit{nf}$=64)  & 7B & 87.67 & 79.50 & 92.61 & 93.50 & 91.50 & 91.00 & 100.00 & 62.90 \\
EVA ($\mathit{mt}$=16, $\mathit{nf}$=128) & 7B & 87.83 & 82.50 & 94.32 & 91.50 & 92.50 & 91.50 &  97.00 & 60.48 \\
\bottomrule
\end{tabularx}
\label{tab:videohallucer_basic}
\caption{VideoHallucer \citep{videohallucer2024} \textbf{Basic Accuracy} (\%) per category.
Obj-Rel = Object Relation, Sem-Det = Semantic Detail, Ext-Fact = External Factual, Ext-NF = External Non-Factual, Fact-Det = Fact Detection, Interact = Interaction.}
\end{table*}

\begin{table*}[t]
\centering
\scriptsize
\setlength{\tabcolsep}{4pt}
\begin{tabularx}{\linewidth}{l|c|c|ccccccc}
\toprule
\textbf{Method} & \textbf{Size} & \textbf{Overall} & \textbf{Obj-Rel} & \textbf{Temporal} & \textbf{Sem-Det} & \textbf{Ext-Fact} & \textbf{Ext-NF} & \textbf{Fact-Det} & \textbf{Interact} \\
\midrule
Video-R1 ($\mathit{nf}$=16) & 7B & 55.50 & 68.50 & 90.34 & 63.50 & 14.00 & 55.50 & 50.00 & 43.55 \\
Video-R1 ($\mathit{nf}$=32) & 7B & 45.50 & 24.00 & 86.36 & 54.50 & 14.00 & 55.00 & 49.00 & 40.32 \\
Video-R1 ($\mathit{nf}$=64) & 7B & 41.08 &  5.00 & 85.23 & 37.50 & 14.50 & 54.00 & 64.00 & 45.97 \\
\midrule
VideoMind ($\mathit{nf}$=16) & 7B & 63.75 & 76.50 & 88.64 & 75.00 & 26.00 & 63.50 & 41.00 & 69.35 \\
VideoMind ($\mathit{nf}$=32) & 7B & 64.58 & 76.50 & 88.64 & 74.50 & 33.00 & 67.00 & 43.00 & 59.68 \\
VideoMind ($\mathit{nf}$=64) & 7B & 62.33 & 77.00 & 88.64 & 71.50 & 28.00 & 66.00 & 41.00 & 53.23 \\
\midrule
EVA ($\mathit{mt}$=4,  $\mathit{nf}$=16)  & 7B & 46.00 & 72.50 & 43.75 & 59.00 & 24.50 & 52.00 & 28.00 & 25.00 \\
EVA ($\mathit{mt}$=4,  $\mathit{nf}$=32)  & 7B & 45.83 & 72.50 & 42.61 & 56.00 & 20.00 & 53.00 & 40.00 & 25.81 \\
EVA ($\mathit{mt}$=4,  $\mathit{nf}$=64)  & 7B & 45.75 & 73.50 & 37.50 & 58.50 & 22.00 & 56.50 & 30.00 & 25.81 \\
EVA ($\mathit{mt}$=4,  $\mathit{nf}$=128) & 7B & 46.17 & 74.50 & 43.75 & 59.00 & 19.00 & 55.50 & 31.00 & 24.19 \\
\midrule
EVA ($\mathit{mt}$=6,  $\mathit{nf}$=16)  & 7B & 46.92 & 73.50 & 44.32 & 57.00 & 22.50 & 50.50 & 36.00 & 33.87 \\
EVA ($\mathit{mt}$=6,  $\mathit{nf}$=32)  & 7B & 46.17 & 72.50 & 44.32 & 58.00 & 20.50 & 51.00 & 37.00 & 28.23 \\
EVA ($\mathit{mt}$=6,  $\mathit{nf}$=64)  & 7B & 45.50 & 74.50 & 42.05 & 58.00 & 18.50 & 52.50 & 31.00 & 27.42 \\
EVA ($\mathit{mt}$=6,  $\mathit{nf}$=128) & 7B & 46.25 & 75.50 & 40.91 & 57.00 & 22.00 & 51.50 & 35.00 & 29.03 \\
\midrule
EVA ($\mathit{mt}$=8,  $\mathit{nf}$=16)  & 7B & 46.33 & 75.50 & 44.32 & 59.50 & 20.00 & 50.50 & 32.00 & 28.23 \\
EVA ($\mathit{mt}$=8,  $\mathit{nf}$=32)  & 7B & 46.67 & 70.00 & 44.32 & 57.50 & 24.00 & 55.50 & 34.00 & 27.42 \\
EVA ($\mathit{mt}$=8,  $\mathit{nf}$=64)  & 7B & 45.50 & 75.00 & 39.20 & 59.50 & 19.00 & 53.00 & 34.00 & 24.19 \\
EVA ($\mathit{mt}$=8,  $\mathit{nf}$=128) & 7B & 46.33 & 75.00 & 44.89 & 56.00 & 19.50 & 55.50 & 28.00 & 29.84 \\
\midrule
EVA ($\mathit{mt}$=10, $\mathit{nf}$=16)  & 7B & 46.33 & 73.50 & 43.75 & 59.00 & 22.00 & 49.00 & 34.00 & 30.65 \\
EVA ($\mathit{mt}$=10, $\mathit{nf}$=32)  & 7B & 46.17 & 76.00 & 40.34 & 59.50 & 22.50 & 48.50 & 40.00 & 24.19 \\
EVA ($\mathit{mt}$=10, $\mathit{nf}$=64)  & 7B & 47.17 & 75.00 & 43.75 & 60.00 & 20.50 & 55.50 & 30.00 & 29.84 \\
EVA ($\mathit{mt}$=10, $\mathit{nf}$=128) & 7B & 47.33 & 74.50 & 45.45 & 59.50 & 19.00 & 54.00 & 36.00 & 30.65 \\
\midrule
EVA ($\mathit{mt}$=16, $\mathit{nf}$=16)  & 7B & 45.58 & 72.00 & 43.18 & 55.50 & 20.50 & 51.00 & 32.00 & 33.06 \\
EVA ($\mathit{mt}$=16, $\mathit{nf}$=32)  & 7B & 46.67 & 74.00 & 43.75 & 59.00 & 21.50 & 52.00 & 40.00 & 24.19 \\
EVA ($\mathit{mt}$=16, $\mathit{nf}$=64)  & 7B & 47.33 & 72.00 & 42.61 & 59.00 & 22.00 & 57.00 & 37.00 & 29.03 \\
EVA ($\mathit{mt}$=16, $\mathit{nf}$=128) & 7B & 47.25 & 75.50 & 43.18 & 58.50 & 19.00 & 54.50 & 38.00 & 30.65 \\
\bottomrule
\end{tabularx}
\label{tab:videohallucer_halluc}
\caption{VideoHallucer \citep{videohallucer2024} \textbf{Hallucination Accuracy} (\%) per category.}

\end{table*}

\begin{table*}[t]
\centering
\scriptsize
\setlength{\tabcolsep}{4pt}

\begin{tabularx}{\linewidth}{l|c|c|ccccccc}
\toprule
\textbf{Method} & \textbf{Size} & \textbf{Overall} & \textbf{Obj-Rel} & \textbf{Temporal} & \textbf{Sem-Det} & \textbf{Ext-Fact} & \textbf{Ext-NF} & \textbf{Fact-Det} & \textbf{Interact} \\
\midrule
Video-R1 ($\mathit{nf}$=16) & 7B & 42.58 & 52.00 & 48.86 & 58.00 & 12.00 & 48.00 & 49.00 & 29.03 \\
Video-R1 ($\mathit{nf}$=32) & 7B & 34.08 & 20.00 & 43.18 & 46.50 & 12.50 & 49.50 & 47.00 & 23.39 \\
Video-R1 ($\mathit{nf}$=64) & 7B & 29.17 &  3.50 & 51.14 & 19.00 & 11.50 & 47.50 & 59.00 & 30.65 \\
\midrule
VideoMind ($\mathit{nf}$=16) & 7B & 44.58 & 61.00 & 46.59 & 64.50 & 21.00 & 54.50 & 11.00 & 32.26 \\
VideoMind ($\mathit{nf}$=32) & 7B & 46.83 & 61.50 & 49.43 & 67.00 & 27.50 & 58.00 & 13.00 & 27.42 \\
VideoMind ($\mathit{nf}$=64) & 7B & 46.58 & 62.50 & 53.98 & 60.50 & 24.00 & 59.50 & 22.00 & 23.39 \\
\midrule
EVA ($\mathit{mt}$=4,  $\mathit{nf}$=16)  & 7B & 40.67 & 61.00 & 42.05 & 52.50 & 21.00 & 47.50 & 28.00 & 17.74 \\
EVA ($\mathit{mt}$=4,  $\mathit{nf}$=32)  & 7B & 39.08 & 60.00 & 42.05 & 50.00 & 17.00 & 44.00 & 39.00 & 11.29 \\
EVA ($\mathit{mt}$=4,  $\mathit{nf}$=64)  & 7B & 39.67 & 61.00 & 36.93 & 53.50 & 20.00 & 49.00 & 30.00 & 11.29 \\
EVA ($\mathit{mt}$=4,  $\mathit{nf}$=128) & 7B & 39.42 & 59.00 & 41.48 & 54.50 & 16.00 & 47.50 & 29.00 & 13.71 \\
\midrule
EVA ($\mathit{mt}$=6,  $\mathit{nf}$=16)  & 7B & 40.42 & 60.50 & 43.75 & 51.50 & 19.50 & 45.50 & 35.00 & 15.32 \\
EVA ($\mathit{mt}$=6,  $\mathit{nf}$=32)  & 7B & 39.50 & 59.00 & 42.61 & 52.00 & 16.50 & 43.00 & 37.00 & 16.94 \\
EVA ($\mathit{mt}$=6,  $\mathit{nf}$=64)  & 7B & 39.17 & 61.50 & 40.91 & 52.50 & 14.50 & 45.50 & 31.00 & 15.32 \\
EVA ($\mathit{mt}$=6,  $\mathit{nf}$=128) & 7B & 39.00 & 59.00 & 39.77 & 52.00 & 17.50 & 44.50 & 35.00 & 13.71 \\
\midrule
EVA ($\mathit{mt}$=8,  $\mathit{nf}$=16)  & 7B & 39.75 & 62.00 & 41.48 & 54.00 & 16.50 & 43.00 & 32.00 & 16.94 \\
EVA ($\mathit{mt}$=8,  $\mathit{nf}$=32)  & 7B & 40.50 & 59.00 & 44.32 & 51.50 & 19.00 & 49.00 & 34.00 & 13.71 \\
EVA ($\mathit{mt}$=8,  $\mathit{nf}$=64)  & 7B & 39.42 & 61.00 & 37.50 & 54.50 & 15.50 & 48.00 & 34.00 & 12.10 \\
EVA ($\mathit{mt}$=8,  $\mathit{nf}$=128) & 7B & 40.08 & 62.00 & 43.18 & 52.00 & 16.00 & 49.00 & 28.00 & 15.32 \\
\midrule
EVA ($\mathit{mt}$=10, $\mathit{nf}$=16)  & 7B & 40.00 & 60.00 & 41.48 & 54.50 & 18.00 & 42.50 & 34.00 & 18.55 \\
EVA ($\mathit{mt}$=10, $\mathit{nf}$=32)  & 7B & 39.92 & 62.00 & 40.34 & 54.50 & 18.00 & 42.00 & 40.00 & 12.10 \\
EVA ($\mathit{mt}$=10, $\mathit{nf}$=64)  & 7B & 39.75 & 60.00 & 42.05 & 54.50 & 17.00 & 47.50 & 30.00 & 12.10 \\
EVA ($\mathit{mt}$=10, $\mathit{nf}$=128) & 7B & 40.67 & 60.50 & 43.75 & 54.50 & 14.00 & 46.50 & 35.00 & 20.16 \\
\midrule
EVA ($\mathit{mt}$=16, $\mathit{nf}$=16)  & 7B & 39.17 & 56.50 & 41.48 & 51.00 & 18.00 & 45.50 & 31.00 & 19.35 \\
EVA ($\mathit{mt}$=16, $\mathit{nf}$=32)  & 7B & 40.08 & 57.00 & 43.75 & 53.00 & 18.00 & 45.00 & 40.00 & 14.52 \\
EVA ($\mathit{mt}$=16, $\mathit{nf}$=64)  & 7B & 40.58 & 58.00 & 41.48 & 54.50 & 17.50 & 49.00 & 37.00 & 15.32 \\
EVA ($\mathit{mt}$=16, $\mathit{nf}$=128) & 7B & 40.17 & 62.00 & 42.05 & 52.50 & 15.50 & 47.00 & 36.00 & 14.52 \\
\bottomrule
\end{tabularx}
\label{tab:videohallucer_both}
\caption{VideoHallucer \citep{videohallucer2024} \textbf{Both Accuracy} (\%) per category --- both basic and hallucination questions answered correctly.}

\end{table*}


\end{document}